\pdfoutput=1

\documentclass[11pt]{article}

\usepackage[final]{acl}
 
\usepackage{times}
\usepackage{latexsym}
\usepackage{amsmath}
\usepackage{amsfonts}
\usepackage{algorithm}
\usepackage{algpseudocode}
\usepackage{booktabs}
\usepackage{adjustbox}

\usepackage{subcaption}
\usepackage{tabularx}
\usepackage{hyperref}

\usepackage[T1]{fontenc}

\usepackage[utf8]{inputenc}

\usepackage{microtype}

\usepackage{inconsolata}

\usepackage{graphicx}

\usepackage[normalem]{ulem}

\newcommand{\doc}{\mathrm{doc}}
\newcommand{\sent}{\mathrm{sent}}
\newcommand{\ai}{\mathrm{AI}}
\newcommand{\human}{\mathrm{Human}}
\newcommand{\mixed}{\mathrm{Mixed}}
\newcommand{\agg}{\mathrm{agg}}
\newcommand{\concat}{\mathrm{concat}}

\newcommand{\gptzero}{GPTZero}

\newcommand{\hellosimple}{HC3}

\newcommand{\mixednormal}{\texttt{with mixed}}
\newcommand{\mixedasai}{\texttt{mixed labeled ai}}
\newcommand{\mixedasmajority}{\texttt{mixed labeled majority}}

\DeclareMathOperator*{\argmax}{arg\,max}

\title{\gptzero: Robust Detection of LLM-Generated Texts}

\author{
 \textbf{George Alexandru Adam}*\textsuperscript{1},
 \textbf{Alexander Cui}*\textsuperscript{1},
 \textbf{Edwin Thomas}\textsuperscript{1},
 \textbf{Emily Napier}\textsuperscript{1},
 \textbf{Nazar Shmatko}\textsuperscript{1},
 \textbf{Yue Gao}\textsuperscript{1},
\\
 \textbf{Zikun Chen}\textsuperscript{1},
 \textbf{Jacob Schnell}\textsuperscript{3,\dag},
 \textbf{Jacob Junqi Tian}\textsuperscript{4,5,\dag},
 \textbf{Alekhya Dronavalli}\textsuperscript{\dag},
 \textbf{Edward Tian}\textsuperscript{1},
 \textbf{Dongwon Lee}\textsuperscript{1,2}
\\
\\
 \textsuperscript{1}GPTZero,
 \textsuperscript{2}Pennsylvania State University
 \textsuperscript{3}University of Waterloo
 \textsuperscript{4}Vector Institute
 \textsuperscript{5}Mila
\\
 \small{
   \textbf{Correspondence:} \href{mailto:email@domain}{alex.adam@gptzero.me}
 }
}

\begin{document}
\maketitle
\begin{abstract}
While historical considerations surrounding text authenticity revolved primarily around plagiarism, the advent of large language models (LLMs) has introduced a new challenge: distinguishing human-authored from AI-generated text. 
This shift raises significant concerns, including the undermining of skill evaluations, the mass-production of low-quality content, and the proliferation of misinformation. 
Addressing these issues, we introduce {\bf {\gptzero} 5.0}, 
a state-of-the-art industrial AI detection solution, offering reliable discernment between human and LLM-generated text. 
Our key contributions include: introducing a hierarchical, multi-task architecture enabling a flexible taxonomy of human and AI texts, setting a new state-of-the-art for AI detection accuracy on 11 independent third-party benchmarks and 18 writing domains, and achieving superior robustness to adversarial attacks and paraphrasing via multi-tiered automated red teaming.
\gptzero{} offers accurate and explainable detection, and educates users on its responsible use, ensuring fair and transparent assessment of text.
\end{abstract}

\begingroup
\renewcommand{\thefootnote}{\fnsymbol{footnote}}
\footnotetext[1]{Denotes equal contribution.}
\footnotetext[2]{Work done while at \gptzero.}
\endgroup

\section{Introduction}
Text is the most ubiquitous medium for representing and communicating information, being used for millennia to circulate ideas in academic, business, and entertainment settings. 
The primary concern on the authenticity of human-written text has typically been plagiarism \cite{Randall2001-dw}, but the widespread use of large language models (LLMs) now raises a question of whether a text is truly human-authored or written by an algorithm \cite{Padillah2024-js}.


Among the issues posed by the widespread access to AI text-generators is the undermining of assessments, where individuals may utilize LLMs to misrepresent their skills and knowledge, resulting in worthless certifications and unfounded hiring decisions.
The proliferation of unintentionally AI-generated content in training corpora has a harmful effect on LLM performance \cite{penedo2024finewebdatasetsdecantingweb}.
Lastly, AI text generators pose significant societal harm, whereby malicious actors can spread misinformation \cite{Yang2024-pz, maung24} or overwhelm academic journals with excessive low-quality submissions \cite{Liang2024-iu, Latona2024-lw}.

\begin{table*}[!tp]
\centering
\scriptsize
\setlength{\tabcolsep}{3.5pt}
\begin{adjustbox}{max width=\linewidth, max totalheight=0.85\textheight}
\begin{tabular}{llcrrrrrrrr}
\toprule
 & & & & \multicolumn{2}{c}{GPTZero} & & \multicolumn{4}{c}{Open-source baselines} \\
\cmidrule(lr){5-6}\cmidrule(lr){8-11}
Benchmark & Subset & Metric & $n$ & 5.0b & 4.11b & Pangram 4 & HC3 & RADAR & Binoculars & Fast-DetectGPT \\
\midrule
Peer Review & Human reviews & FPR$\downarrow$ & 3,498 & \textbf{0.00} & 0.06 & \textbf{0.00}$^{\S}$ & 0.09 & 21.44 & 2.77 & 19.13 \\
 & AI, basic prompt & Recall$\uparrow$ & 6,121 & \textbf{99.82} & 99.41 & 97.80$^{\S}$ & 12.45 & 37.40 & 75.69 & 86.36 \\
 & AI, enhanced prompt & Recall$\uparrow$ & 5,332 & \textbf{99.81} & 99.55 & 98.50$^{\S}$ & 14.70 & 38.18 & 77.70 & 86.76 \\
 & Humanized AI & Recall$\uparrow$ & 1,000 & 98.20 & \textbf{98.90} & 78.50$^{\S}$ & 0.00 & 41.30 & 0.20 & 2.40 \\
 & AI-edited human$^{\ast}$ & AI verdict$\uparrow$ & 16,046 & \textbf{96.96} & 96.25 & 58.70$^{\S}$ & 0.89 & 26.86 & 17.74 & 48.83 \\
 & Humanized AI-edited$^{\ast}$ & AI verdict$\uparrow$ & 1,000 & \textbf{90.20} & 88.50 & 71.70$^{\S}$ & 0.10 & 43.10 & 0.00 & 0.00 \\
\midrule
Sem-Detect & Human vs.\ rest & AUC$\uparrow$ & 4,454 & \textbf{$>$99.99} & \textbf{$>$99.99} & 97.80 & 77.64 & 30.27 & 75.78 & 68.20 \\
 & AI-generated & TPR@0.1\%$\uparrow$ & 1,372 & \textbf{100.00} & \textbf{100.00} & 95.50 & 1.31 & 0.00 & 0.07 & 0.00 \\
 & AI-generated & TPR@1\%$\uparrow$ & 1,372 & \textbf{100.00} & \textbf{100.00} & 95.50 & 6.41 & 0.00 & 2.62 & 0.58 \\
\midrule
DetectRL & Multi-domain F1 & F1$\uparrow$ & 48,586 & \textbf{98.87} & 98.45 & 96.93 & 67.44 & 85.28 & 77.80 & 64.40 \\
 & Multi-LLM F1 & F1$\uparrow$ & 48,586 & \textbf{99.06} & 98.38 & 97.25 & 74.43 & 84.48 & 75.84 & 60.32 \\
 & Multi-attack F1 & F1$\uparrow$ & 48,586 & \textbf{97.15} & 96.69 & 96.00 & 75.46 & 83.50 & 78.48 & 67.94 \\
 & Domain generalization F1 & F1$\uparrow$ & 48,586 & \textbf{98.71} & 97.95 & 96.95 & 55.14 & 82.42 & 76.91 & 64.16 \\
 & LLM generalization F1 & F1$\uparrow$ & 48,586 & \textbf{98.24} & 96.90 & 97.25 & 69.83 & 82.53 & 76.39 & 63.70 \\
 & Attack generalization F1 & F1$\uparrow$ & 48,586 & \textbf{96.31} & 96.10 & 95.99 & 81.99 & 85.76 & 79.14 & 68.41 \\
 & Time, test F1 & F1$\uparrow$ & 48,586 & \textbf{94.54} & 92.12 & 91.30 & 62.93 & 64.20 & 74.57 & 59.46 \\
 & Human writing F1 & F1$\uparrow$ & 48,586 & \textbf{95.82} & 95.63 & 90.77 & 79.39 & 84.95 & 90.33 & 82.47 \\
 & Average F1 & F1$\uparrow$ & 48,586 & \textbf{97.34} & 96.53 & 95.30 & 70.83 & 81.64 & 78.68 & 66.36 \\
\midrule
UChicago & Human & FPR$\downarrow$ & 1,992 & \textbf{0.00} & \textbf{0.00} & \textbf{0.00} & 10.54 & 22.34 & 4.07 & 18.98 \\
 & Standard, full length & TPR@1\%$\uparrow$ & 7,310 & \textbf{100.00} & 99.99 & \textbf{100.00} & 0.59 & 0.00 & 43.75 & 51.48 \\
 & Humanized, full length & TPR@1\%$\uparrow$ & 7,635 & 95.99 & 95.53 & \textbf{98.93} & 0.28 & 0.01 & 0.83 & 10.31 \\
 & Standard, $<$50 words & TPR@1\%$\uparrow$ & 658 & \textbf{99.70} & 98.02 & \textbf{99.70} & 4.10 & 0.91 & 4.86 & 0.00 \\
 & Humanized, $<$50 words & TPR@1\%$\uparrow$ & 331$^{\dagger}$ & \textbf{83.08} & 63.44 & 73.32 & 2.72 & 3.63 & 4.53 & 0.00 \\
\midrule
GEDE & Overall & TPR@1\%$\uparrow$ & 474$^{\dagger}$ & \textbf{95.57} & 91.77 & 94.94$^{\P}$ & 20.89 & 20.25 & 78.69 & 54.43 \\
 & Fully generated & TPR@1\%$\uparrow$ & 190$^{\dagger}$ & \textbf{100.00} & \textbf{100.00} & \textbf{100.00}$^{\P}$ & 28.95 & 31.58 & \textbf{100.00} & 84.74 \\
 & AI-improved human & TPR@1\%$\uparrow$ & 190$^{\dagger}$ & \textbf{88.95} & 79.47 & \textbf{88.95}$^{\P}$ & 7.89 & 7.89 & 48.42 & 10.53 \\
 & Humanized & TPR@1\%$\uparrow$ & 94$^{\dagger}$ & \textbf{100.00} & \textbf{100.00} & 96.81$^{\P}$ & 30.85 & 22.34 & 96.81 & 81.91 \\
\midrule
Epoch AI & AI & FNR$\downarrow$ & 594 & \textbf{1.18} & 1.52 & 2.86 & 96.80 & 99.83 & 89.56 & 52.02 \\
 & Human & FPR$\downarrow$ & 495$^{\dagger}$ & 0.20 & 0.20 & \textbf{0.00} & 1.41 & 0.61 & \textbf{0.00} & 23.64 \\
\midrule
GRiD & AI & TPR@1\%$\uparrow$ & 1,368 & \textbf{99.93} & 97.81 & 99.63$^{\P}$ & 48.46 & 0.29 & 86.18 & 54.17 \\
 & All texts & F1@1\%$\uparrow$ & 6,513 & \textbf{99.38} & 97.20 & 98.31$^{\P}$ & 63.69 & 0.57 & 90.76 & 68.67 \\
\midrule
Acad.\ Integrity & Original abstracts & FPR$\downarrow$ & 306$^{\dagger}$ & \textbf{0.00} & 0.33 & 0.30$^{\S}$ & 5.23 & 24.84 & 8.17 & 20.26 \\
 & Refined/regenerated$^{\ast}$ & AI verdict$\uparrow$ & 918 & \textbf{97.06} & 95.86 & 96.40$^{\S}$ & 4.47 & 14.71 & 15.90 & 41.18 \\
\midrule
MELD & Human & FPR$\downarrow$ & 7,862 & \textbf{0.01} & 0.11 & -- & 16.74 & 3.05 & 3.04 & 20.13 \\
 & Generator: GPT-5.4-Mini & TPR@1\%$\uparrow$ & 7,862 & \textbf{100.00} & 99.95 & 99.99 & 2.53 & 1.28 & 37.60 & 25.22 \\
 & Generator: Gemini-3-Flash & TPR@1\%$\uparrow$ & 7,862 & \textbf{100.00} & 99.95 & 99.99 & 4.59 & 3.08 & 56.60 & 37.69 \\
 & Generator: Claude-Haiku-4.5 & TPR@1\%$\uparrow$ & 7,862 & \textbf{100.00} & 99.90 & 99.99 & 3.38 & 5.83 & 75.11 & 26.99 \\
 & Generator: Qwen-3.6-Plus & TPR@1\%$\uparrow$ & 7,862 & \textbf{100.00} & 99.97 & 99.99 & 5.16 & 2.42 & 75.62 & 47.16 \\
 & Generator: Overall & TPR@1\%$\uparrow$ & 31,448 & \textbf{100.00} & 99.94 & 99.99 & 3.92 & 3.15 & 61.23 & 34.27 \\
 & Attack: No attack & TPR@1\%$\uparrow$ & 31,448 & \textbf{100.00} & 99.94 & \textbf{100.00} & 3.92 & 3.15 & 61.23 & 34.27 \\
 & Attack: Homoglyph & TPR@1\%$\uparrow$ & 31,448 & \textbf{100.00} & 99.94 & 99.97 & 0.00 & 0.06 & 16.98 & 20.99 \\
 & Attack: Number perturbation & TPR@1\%$\uparrow$ & 31,448 & \textbf{100.00} & 99.94 & \textbf{100.00} & 3.64 & 3.08 & 53.04 & 31.51 \\
 & Attack: Synonym & TPR@1\%$\uparrow$ & 31,448 & \textbf{100.00} & 99.94 & 99.96 & 1.99 & 2.04 & 23.60 & 11.06 \\
 & Attack: Upper--lower flip & TPR@1\%$\uparrow$ & 31,448 & \textbf{100.00} & 99.96 & 99.99 & 0.00 & 0.01 & 0.87 & 4.82 \\
 & Attack: Whitespace & TPR@1\%$\uparrow$ & 31,448 & \textbf{100.00} & 99.94 & 99.99 & 0.00 & 1.89 & 29.87 & 18.07 \\
 & Attack: Zero-width space & TPR@1\%$\uparrow$ & 31,448 & \textbf{100.00} & 99.94 & \textbf{100.00} & 0.00 & 1.46 & 36.83 & 15.66 \\
 & Attack: Overall & TPR@1\%$\uparrow$ & 220,136 & \textbf{100.00} & 99.94 & 99.99 & 1.37 & 1.67 & 31.77 & 19.48 \\
\midrule
Russell & Human & FPR$\downarrow$ & 150$^{\dagger}$ & 0.67 & \textbf{0.00} & 0.67 & 6.00 & 2.67 & 0.67 & 26.67 \\
 & GPT-4o & Recall$\uparrow$ & 30$^{\dagger}$ & \textbf{100.00} & \textbf{100.00} & \textbf{100.00} & \textbf{100.00} & 83.33 & \textbf{100.00} & \textbf{100.00} \\
 & Claude & Recall$\uparrow$ & 30$^{\dagger}$ & \textbf{100.00} & \textbf{100.00} & \textbf{100.00} & 60.00 & 0.00 & 96.67 & \textbf{100.00} \\
 & o1-pro & Recall$\uparrow$ & 30$^{\dagger}$ & \textbf{100.00} & \textbf{100.00} & \textbf{100.00} & 0.00 & 0.00 & 66.67 & \textbf{100.00} \\
 & Paraphrase evasion & Recall$\uparrow$ & 30$^{\dagger}$ & \textbf{100.00} & \textbf{100.00} & \textbf{100.00} & 30.00 & 40.00 & 63.33 & 83.33 \\
 & Humanize evasion & Recall$\uparrow$ & 30$^{\dagger}$ & \textbf{100.00} & \textbf{100.00} & \textbf{100.00} & 0.00 & 0.00 & 0.00 & 26.67 \\
\midrule
VUB & Deep Research papers & Score $\geq$80\%$\uparrow$ & 39$^{\dagger}$ & \textbf{100.00} & 94.87 & \textbf{100.00} & 0.00 & 0.00 & n/a & 0.00 \\
\bottomrule
\end{tabular}
\end{adjustbox}
\par\vspace{2pt}
\parbox{\linewidth}{\scriptsize All values in \%; best value per row in bold, with ties at two decimals. $\uparrow$/$\downarrow$: higher/lower is better. $n$: samples per row for GPTZero and the open-source baselines; $^{\dagger}$$n<500$. FPR, FNR, recall and AI-verdict rows use each detector's own decision (baselines: $P(\mathrm{AI})\geq 0.5$; Binoculars: score $<0.9015$). TPR@$x$\% thresholds are set on each benchmark's human texts; F1@1\% is F1 over all texts at the TPR@1\% threshold with the fewest false positives. DetectRL follows its leaderboard protocol. $^{\ast}$AI-verdict rate on human text edited by an LLM. Pangram 4 values are as reported (--: not reported; $^{\S}$on a subset of the samples), except $^{\P}$, which we obtained by scoring every sample through the Pangram 4 API. $>$99.99: rounds to 100.00 but is not exactly 100. n/a: Binoculars outputs no probability.}
\caption{Results of GPTZero 5.0b and 4.11b, Pangram 4 and four open-source baselines on 11 third-party benchmarks: Peer Review~\cite{saha2026peerreview}, Sem-Detect~\cite{pmlr-v306-duarte26a}, DetectRL~\cite{NEURIPS2024_b61bdf7e}, UChicago~\cite{jabarian2025artificial}, GEDE~\cite{gehring2025gede}, Epoch AI~\cite{lee2026epochdetectors}, GRiD~\cite{Qazi_2024}, Acad.\ Integrity~\cite{karr2026academic}, MELD~\cite{li2026meld}, Russell~\cite{russell2025human}, VUB~\cite{vanvlasselaer2026who}.}
\label{tab:third-party}
\end{table*}


\begin{figure}
\centering
\includegraphics[scale=0.4]{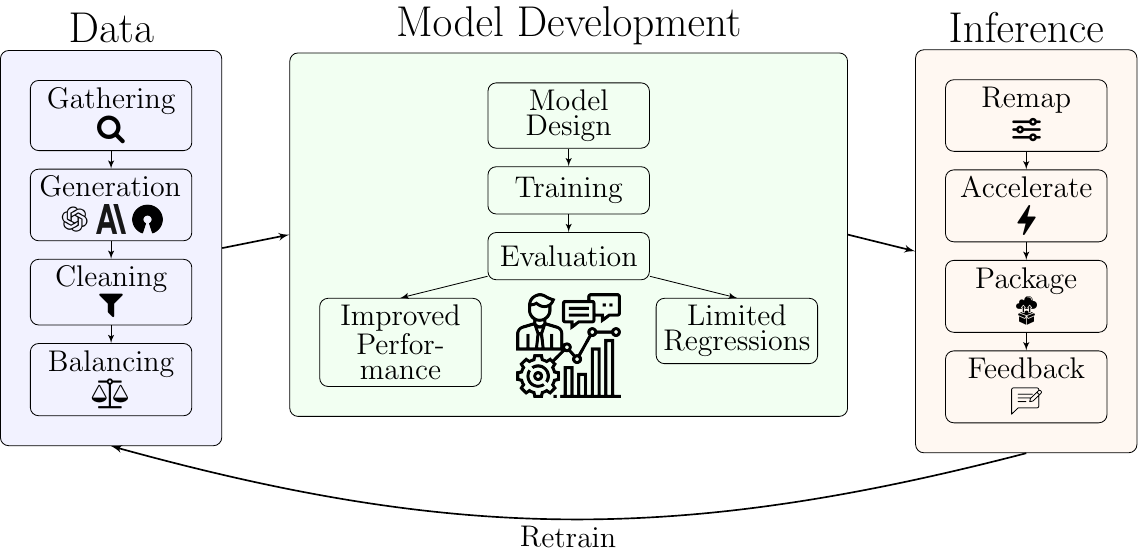}
\caption{The process of developing and improving our deep learning model for detecting LLM-generated texts.
}
\label{figure:overall}
\end{figure}

\gptzero{}'s mission is to ensure that human-written and LLM-generated text can be distinguished by anyone. 
We achieve this by offering an AI detector that is highly accurate, robust, and explainable. Existing commercial solutions are opaque, and do not explain the sentences and corresponding features driving a given prediction. This leaves users confused and distrustful of the black-box predictions. We address this limitation through our features focused on interpretability.
An overview of our approach is shown in Figure \ref{figure:overall}. 

Our contributions are:
    (1) We present a novel use of a hierarchical, multi-task classification architecture for AI detection enabling a flexible, fine-grained taxonomy of human texts, AI texts, and combinations thereof. 
    (2) We demonstrate superior robustness to adversarial attacks and paraphrasers achieved via a multi-tiered red teaming approach
    (3) We present state-of-the-art detection results on 11 independent third-party benchmarks, 18 writing domains, 50 generating LLMs including the newest frontier models, compared with open- and closed-source competitors.
    (4) We report false-positive rates for English language learners and for student demographic groups, where our latest model makes no false positive in any group.

\section{Related Work}
\label{sec:related}

{\bf Statistical/Metric Based Detection}. A principled way of testing if a text was generated by an LLM is to evaluate the likelihood of that text under a statistical model which serves as a proxy \cite{Su2023-de}. 
Several variations on this approach have emerged, including computing the average rank of the text's words in the output distribution of an LLM \cite{Su2023-de}, computing the perplexity of the LLM on the text, or even token cohesiveness \cite{Ma2024-wz}.
GPT-who takes a similar approach by deriving multiple likelihood-based features capturing surprisal \cite{Venkatraman2023-pe}. 
DetectGPT proposes repeatedly replacing words in a text with LLM-suggested alternatives \cite{Mitchell2023-qb}, hypothesizing that these perturbations will more significantly change an LLM's log-probs for human-written texts than for AI-written texts. 
Fast-DetectGPT improves on this idea by only calling the LLM once and factorizing the perturbation effects to improve computational efficiency \cite{Bao2023-ay}. 
Lastly, Binoculars uses two LLMs to compute a cross-perplexity score based on agreement between LLMs \cite{Hans2024-qb}.

{\bf Training-Based Detection}.
\cite{Islam2023-uj} showed limited success using classical models such as logistic regression, SVMs \cite{Hearst1998-ty}, and neural networks \cite{Hochreiter1997-gp} using word frequency features as inputs. 
Conversely, using embeddings from small language models, like RoBERTa, has proven to be effective at detecting AI-generated tweets \cite{Kumarage2023-db}. 
Unsupervised contrastive learning has also shown promise in disentangling human-authored from AI-generated text, while identifying features which generalize to new LLMs \cite{Bhattacharjee2023-yv, Zhang2024-di}. The positive unlabeled learning framework can also be used to learn from large amounts of unlabeled text \cite{Tian2023-ui}.
Ghostbuster \cite{Verma2023-ti} trains a shallow classifier on hand-crafted features based on LLM log-probs to achieve strong generalization results. 
Even within the same domain, more advanced text generation approaches such as using a finetuned domain-specific LLM can make detection more challenging for both deep learning and statistical approaches \cite{Dawkins2025-cg}.
The primary limitation of trained AI-generated text detectors is the demanding need for diverse data, without which, generalization to new domains and LLMs is limited \cite{Dugan2024-qu}.
Lastly, deep learning methods have been shown to be effective at making fine-grained predictions at the token-level, capturing human edits of LLM texts \cite{Kadiyala2025-kc, Artemova2024-jh, Pham2025-fe}

\section{The \gptzero{} Detector}


\subsection{Data}



\begin{figure}[!htbp]
\centering
\includegraphics[width=\linewidth]{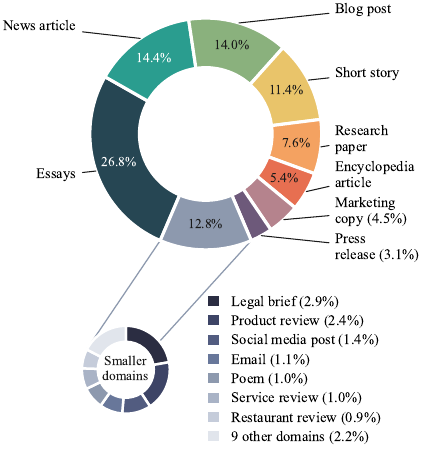}
\caption{Distribution of GPTZero 5.0b's human-written training texts by writing domain. Essay types (expository, argumentative, descriptive, persuasive and admissions essays) are grouped as Essays. The 16 domains that each account for less than 3\% of texts are grouped as Smaller domains and broken down below, all percentages are shares of all human-written training texts.}
\label{fig:training-domains}
\end{figure}

As \gptzero{} uses a deep learning architecture, its performance is proportional to the scale and diversity of the data on which it was trained. The human-written training data of GPTZero 5.0b spans 28 writing domains. Figure~\ref{fig:training-domains} shows
their distribution, and Appendix~\ref{sec:case_study_dataset_details} describes how texts are counted.

\textbf{Data Acquisition}. Our data gathering pipeline, shown in the first step of Figure \ref{figure:overall}, involves regularly collecting publicly available datasets of both human-written and AI-written text.
We additionally use a proprietary generation pipeline using various prompting strategies to generate texts from popular LLM providers such as OpenAI, Anthropic, Google, and open-source LLMs.  


\textbf{Data Cleaning and QA}. Data-driven detectors can be prone to learning biases in the data that help distinguish human and AI text, such as text formatting.
Appendix Table \ref{table:dataset_bias} illustrates some common such biases that must be addressed prior to training to avoid relying on spurious correlations.
We employ both statistical methods and manual inspection datasets to uncover and address issues. 


\textbf{User Feedback Loop}. To help focus our data gathering and generation efforts, we implement a user feedback mechanism where users can dispute predictions they believe to be incorrect (see Appendix Figure \ref{figure:user_feedback}). 
Clustering documents with such feedback enables us to identify the most relevant and underrepresented domains and common failure cases, which is used to scope our future data collection and training efforts.


\subsection{Modeling}
\begin{figure}[h!]
\centering
\includegraphics[width=\linewidth]{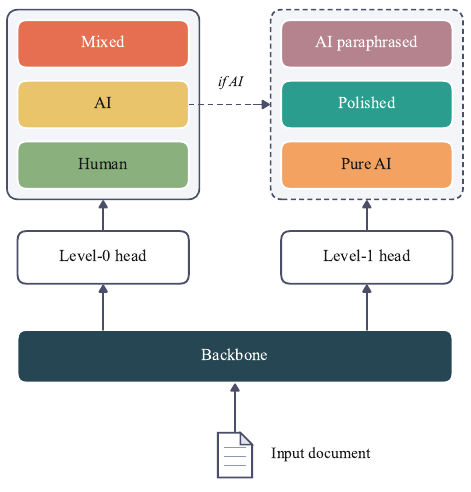}
\caption{Hierarchical Classifier Heads
}
\label{figure:hier_overall}
\end{figure}

The \gptzero{} detector uses a deep learning architecture trained in a supervised fashion, selected for its superior in-distribution performance compared to zero-shot and metric-based models \cite{Lai2024-eh, Tulchinskii2023-mb}. 
A key distinction we make compared to other commercial detectors is in framing AI-generated text detection as a ternary classification problem consisting of Human, AI, and Mixed classes. While this has been framed as a ternary problem in previous research \cite{Zeng2024-qc,Richburg2024-al} , in practice most open sources methods and commercial detectors do not leverage this paradigm.
We use a hierarchical classification head with classes Human, AI, and Mixed at the top level (L0), with finer grained classes Pure AI, Polished (Human written and then AI polished), and AI Paraphrased (AI written and then AI rephrased) under the AI class (L1) as shown in Figure \ref{figure:hier_overall}. 

One advantage this gives us is better sample efficiency for classes with fewer samples such as the AI Paraphrased class, and falling back to the more reliable parent class for low confidence subclass predictions. This novel hierarchical framing of AI detection allows us to communicate to our users the degree to which LLMs were used which is vital in the context of academic integrity. Additionally, the Mixed class allows us to decouple model confidence and the proportion of the text generated by an AI. This resolves the confusing ambiguity of binary classifiers, where a 50\% AI-generated score could either mean the detector is 100\% confident that 50\% of the text is AI-generated, or that the detector is 50\% confident that the document is entirely AI-generated.

By representing a document as a sequence of $n$ sentences $d = (s_{1}, ..., s_{n})$, we enable fine-grained predictions for documents that are a mix of human and AI text.
In particular, \gptzero{} produces both document-level predictions, $f_{\doc}(d) = (p_{\human}, p_{\ai}, p_{\mixed})$, as well as predictions for each sentence individually, $f_{\sent}(d) = (p_{\ai}^{(1)}, ..., p_{\ai}^{(n)})$.
To enable these sentence-level predictions, we adopt a multi-task loss $\mathcal{L} = \mathcal{L}_{d} + \alpha \mathcal{L}_{s}$ where $\mathcal{L}_{d}$ is the document-level cross-entropy loss, $\mathcal{L}_{s}$ is the sentence-level binary cross-entropy loss, and $\alpha$ is a hyperparameter trading-off the two. 
We frame sentence-level predictions as a binary classification problem due to the lack of a well-defined criteria for what mixed sentences are. 
By training our detector with this multi-task objective, \gptzero{} efficiently produces both document and sentence-level predictions in a single forward pass.
These predictions provide significantly finer-grain and more explainable detections over competing detectors discussed in Sec. \ref{sec:related} which only predict a binary document-level prediction, with no principled way of representing mixed documents.
Architecture and hyperparameters are proprietary.

\subsection{Classification of Polished Texts}
\label{sec:polished_method}
We define a Polished text as a human document that has been processed by an LLM to improve grammar, spelling, formatting, clarity, etc. via a prompt such as that in Appendix Section \ref{appendix:polishing_prompts}. Formally, given a human document $d_{\text{Human}}$, a polished text  is defined as $d_{\text{Polished}} = \mathcal{F}_{\text{Polish}}(d_{\text{Human}})$ where $\mathcal{F}_{\text{Polish}}$ is an LLM with a single prompt in its context that has a polishing instruction.

Additionally, we use the following comparison to determine if the text should be included in the dataset of Polished texts or if it should be discarded: $keep = \tau_{\text{min}} \leq sim(d_{\text{Human}}, d_{\text{Polished}}) \leq \tau_{\text{max}}$ where $\text{sim}$ is a similarity metric, $d_{\text{Human}}$ is the human document, $d_{\text{Polished}}$ is the polished document, $\tau_{\text{min}}$ is tuned to ensure that $d_{\text{Polished}}$ is not so different from the original human text that it could be obtained by prompting an LLM without access to $d_{\text{Human}}$ as reference, and $\tau_{\text{max}}$ ensures that sufficient edits were made to $d_{\text{Human}}$ to minimize risk of misclassifying human texts as polished. In practice, we use the Levenshtein Ratio for $sim$ as embedding-based similarity metrics like cosine similarity often fail to capture stylistic differences between texts.

\subsection{Adversarial Robustness}
Many advanced users attempt to undermine AI detectors by editing an LLM-generated text prior to finalizing their document. This process is often referred to as paraphrasing or humanization in the literature \cite{Zhou2024-ud,Zheng2025-lq} and covers a broad set of techniques. Our approach to adversarial robustness involves addressing the following 4 threat models in order of increasing severity: Paraprhrasing Prompts, Paraphrasing Models (e.g. Dipper), Black-Box Humanization Services, White-Box Attacks. We use data augmentation to increase robustness to these threats by transforming the AI texts in our dataset using these techniques.

\begin{figure}
\centering
\includegraphics[width=0.95\columnwidth]{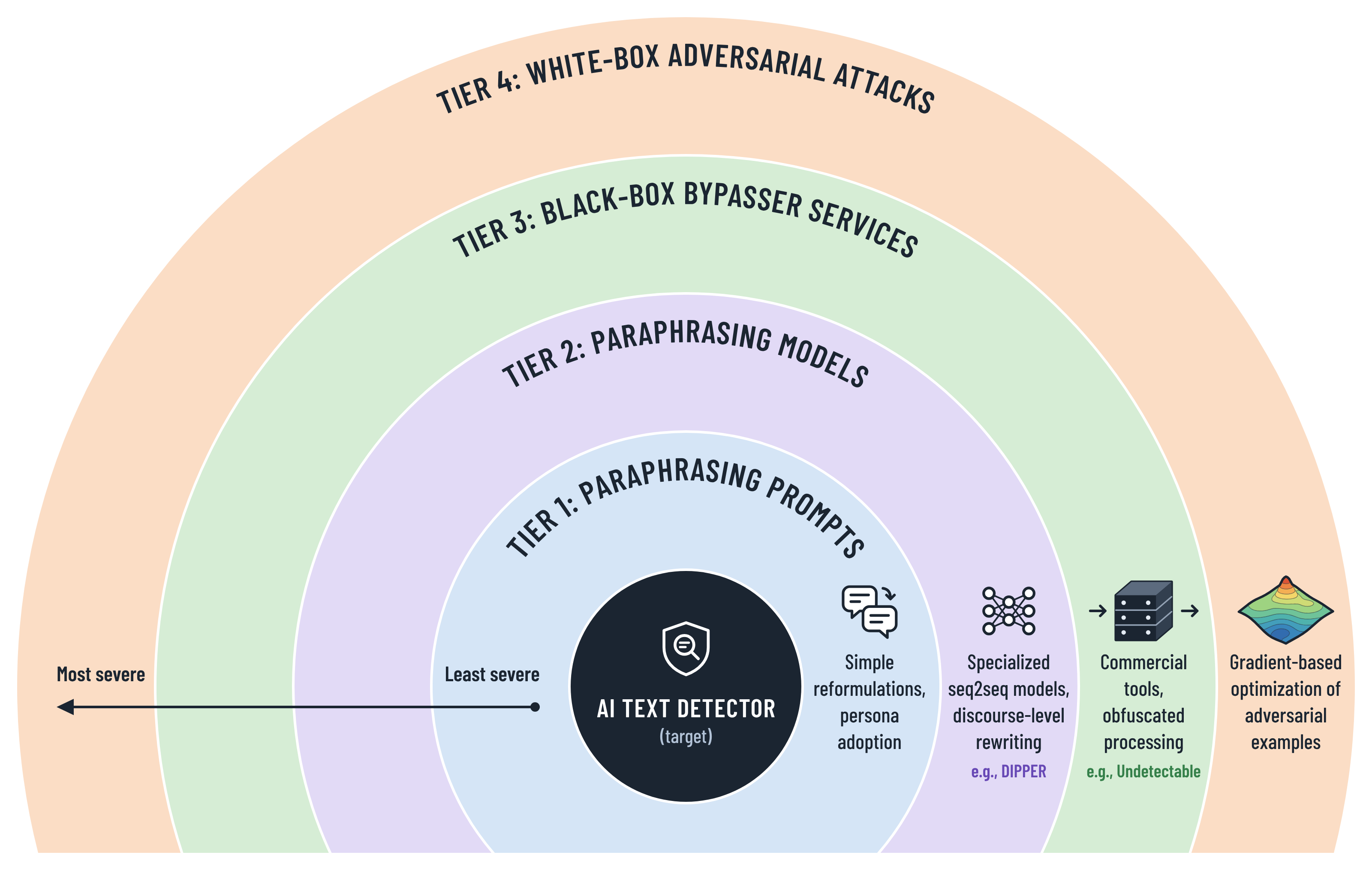}
\caption{GPTZero's multi-tiered red teaming approach covers a variety of adversarial threats, providing unprecedented robustness. 
}
\label{figure:multi_tiered_red_teaming}
\end{figure}

\subsubsection{Paraphrasing Prompts}
The simplest approach to evading AI detection is to prompt the same LLM that generated the text in question to remove vocabulary or punctuation that is common in AI generated text. Inspired by research on LLM watermark removal via translation by \cite{He2024-ox}, we also employ multi-turn prompting strategies, alternating between a paraphrasing prompt, and translation for a chain of up to 4 languages as follows. Let $\mathcal{L} = \{\ell_{1}, \ell_{2}, ..., \ell_{k}\}$ be the sequence of target languages where $k \leq 4$. The following sequence of paraphrased $d_{t}$ is define for $t = 1, ..., 2k$

\begin{align*}
    d_{t} = \begin{cases}
        \mathcal{F}_{\text{Para}}(d_{t-1}) & \text{if } t \bmod 2 = 1 \\
        \mathcal{F}_{\text{Trans}}(d_{t-1}, \ell_{k}) & \!\begin{aligned}[t] 
                                            & \text{if } t \bmod 2 == 0 \\ 
                                            & \text{where } k = t/2 
                                          \end{aligned}
    \end{cases}
\end{align*}

where $\mathcal{F}_{\text{Para}}$ and $\mathcal{F}_{\text{Trans}}$ are a given LLM prompted for paraphrasing and translation respectively.

\subsubsection{Paraphraser Models}
A seminal paraphrasing model called Dipper initially had a high bypass rate against our detector \cite{krishna2023paraphrasing}. Dipper was trained to transform translated versions of English novels back to their English equivalent at the paragraph level. This is a fundamentally different objective than language modeling, resulting in texts that are semantically similar to the original version, but different in vocabulary and sentence structure. Similarly, TempParaphraser uses a paraphrasing model to simulate the randomness that LLMs exhibit for higher temperature settings when sampling, but does so without sacrificing generation quality.

We generate $\alpha N_{AI}$ Dipper texts, and $\beta N_{AI}$ TempParaphraser texts for LLM texts sampled uniformly at random from our training database where $\alpha, \beta \in [0, 1]$. 
In practice $\alpha$ and $\beta$ are proprietary, and vary between model releases.

\subsubsection{Black-Box Paraphraser Services}
Black-box paraphraser services are particularly challenging to defend against since the mechanism for generating the bypassed texts is unknown, allowing for combinations of operations such as: providing few-shot examples of a human writer's style, fine-tuning LLMs for humanization, and even concatenating together semantically similar sentences from a human text database \cite{Chakrabarty2025-ke, Pham2025-fe}. It is impractical to implement all possible approaches, so we instead leverage a limited amount of data from the paraphrasing services listed in Appendix Section \ref{appendix:paraphraser_services}. In particular, we fine-tune our model on these challenging examples along with the white-box attack texts described in the next section.

\subsubsection{White-Box Adversarial Attacks}
The most extreme attack scenario against any machine learning model is white-box adversarial attacks which assume access to model weights and architecture. The discrete nature of text-generation requires approximations to be made as direct gradient-based optimization on a seed text is not tractable. We thus use the approach by \cite{Zhou2024-ud} 
\begin{equation}
\begin{aligned}
& \underset{d_{\text{Adv}}}{\text{minimize}}
& & f_{\text{doc}}(d_{\text{Adv}}) \\
& \text{subject to}
& & d_{\text{Adv}} \in \mathcal{M}(d_{\text{Init}}, \mathcal{R}) \\
& & & C_j(d) = 1, \quad \forall j \in \{1, \dots, K\}
\end{aligned}
\end{equation}

where $f_{\text{doc}}$ is the document-level output for the AI class of our detector, $d_{\text{Init}}$ is the initial document that is being attacked, 
$d_{\text{Adv}}$ is the adversarial document being generated, $\mathcal{M}(d_{\text{Init}}, \mathcal{R})$ is the set of all texts that can be generated by swapping words from $d_{\text{Init}}$ with alternatives provided by the ranker model $\mathcal{R}$. In practice, we use the gradients of our detector to choose the most important input tokens, and perform substitution with  masked language model such as RoBERTa, under the constraint that that perplexity should not increase significantly in order to maintain fluency.

\subsection{Inference}
To ensure \gptzero{}'s robustness and consistency, we apply basic cleaning and reformatting to the document before inference, such as removing extraneous whitespace.
If a document, $d$, does not fit in the detector's context length, we split it into disjoint windows $W_1,\ldots,W_m$ such that $d=\concat(W_1,\ldots,W_m)$ and $|W_i|\le T$ for ${i=1,\ldots,m}$.
We recover a document-level prediction by
$f_{\doc}(d) = \agg(f_{\doc}(W_{1}), ..., f_{\doc}(W_{m}))$
and sentence-level predictions by
$f_{\sent}(d) = \concat(f_{\sent}(W_{1}), ..., f_{\sent}(W_{m}))$,
for some aggregate $\agg$ (e.g., average, median, maximum).

Finally, we leverage a remapping function ${r: \mathbb{R}^3 \rightarrow \mathbb{R}^3}$ to produce our overall document-level predictions $r(f_\doc(d))$.
This post-processing improves calibration and reduces false-positive predictions (Appendix Sec. \ref{sec:calibration}).

\section{Advanced Scan}
\label{sec:appendix_deep_scan}
One of our contributions is providing an unprecedented level of transparency to our users via our feature \textit{Advanced Scan}. \textit{Advanced Scan} is a proprietary method for attributing the document-level prediction of our detector to the sentences that have the largest impact on the prediction (Appendix Figure \ref{fig:deep_scan}). 
Specifically, for each sentence $s_k$ in $d=(s_1,\ldots,s_n)$, Advanced Scan assigns a score $\delta_k$ indicating how the presence of $s_k$ affects $P(y = \mathrm{AI} \,|\, d)$. 
These scores complement the sentence-level probabilities predicted by our detector since those probabilities do not provide the relative importance of each sentence. 
\textit{Advanced Scan} is an adaptation of two methods: Saliency and Occlusion \cite{Atanasova2020-uz}. We use saliency to identify the tokens that would most affect the output of the predicted class if changed. Based on user submissions, we observe that writers mainly replace some words with their synonyms, or remove them altogether. We mimic this behavior to identify how our model's predictions would change in a way that is aligned with user edit behavior. 

\begin{figure}[h!]
\centering
\includegraphics[width=0.9\columnwidth]{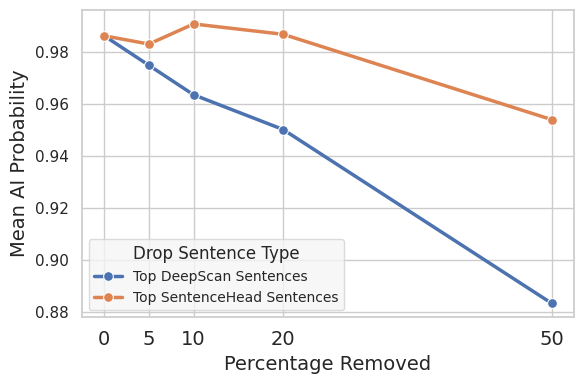}
\caption{Drops in AI probability after removing top-k\% of high AI impact sentences predicted by our Advanced Scan Feature (Blue) and Detector Sentence Head (Orange)
}
\label{figure:adv_scan_faithfulness}
\end{figure}

Figure \ref{figure:adv_scan_faithfulness} shows the drops in AI probability after removing top 5\% to top-50\% of the sentences that have the largest impact on AI scores. The dataset used here consists of 100 documents across a variety of domains. This confirms the faithfulness of Advanced Scan and its ability to identify the most important sentences. 

\section{Case Studies}
In this section, we demonstrate the performance of our detector against commercial and open-source competitors on some cases of interest. 
Data and model outputs are available in our repository.\footnote{
Anonymized Github repository
Placeholder.
}
Dataset details are found in Appendix \ref{sec:case_study_dataset_details}.



\textbf{Evaluation Metrics:} On the third-party benchmarks (Section~\ref{sec:third-party}) we report the metrics each benchmark's authors use. Results at a fixed false-positive rate (TPR@1\% and TPR@0.1\%) are reported using a threshold chosen separately for each detector on the benchmark's own human documents, at their respective best operating points. AUC is threshold-free, and for DetectRL we follow it's leaderboard protocol, which fits a Youden's $J$ threshold per sub-setting and averages F1~\cite{NEURIPS2024_b61bdf7e}. False-positive rate (FPR), false-negative rate (FNR) and AI recall use each detector's own decision: GPTZero's predicted class, Pangram's predicted class, and $P(\mathrm{AI})\geq 0.5$ for the open-source baselines, except Binoculars, where we use its authors' threshold (Appendix~\ref{app:baselines}). We emphasize low false-positive rates, since false positives have a serious negative impact on writers.

\textbf{Closed-Source Comparisons:} In the AI sub-class comparisons below, for competitors with multi-class outputs we sum the probabilities of the non-human classes to obtain an overall AI probability, which is used to compute AUC. Since closed-source competitors present users with a final classification based on internal thresholds, we use that classification as the final prediction for both GPTZero and the closed-source competitors.

We consider the following baselines:
(1) \textbf{Radar} uses a deep learning model trained adversarially \cite{Hu2023-uv},
(2) \textbf{Fast-DetectGPT} is a zero-shot approach for detecting LLM-generated texts based on a heuristic that modified versions of an LLM-generated text should have consistently lower probability than the original version, formalized by using conditional probability curvature \cite{Bao2023-ay},
(3) 
\textbf{Binoculars}
 uses two LLMs to compute a cross-perplexity score which is meant to account for tokens that are unlikely unless a highly specific prompt is used \cite{Hans2024-qb},
(4)
\textbf{\hellosimple} 
 uses a RoBERTa architecture trained on ChatGPT texts and specializes in short text detection \cite{Guo2023-gu}
(5)
\textbf{Pangram (Pangram 4 on the third-party benchmarks, 2.0 for non-English texts)}
 is a commercial detector that prioritizes precision from a plagiarism detection company \cite{Emi2024-nq}. We evaluate GPTZero 5.0b, our latest model, and GPTZero 4.11b, the previous production release, on identical samples. We run the open-source baselines ourselves on every sample of the third-party benchmarks (Appendix~\ref{app:baselines}), while Pangram 4 results are taken as reported.

\subsection{Third-Party Benchmarks}
\label{sec:third-party}

We evaluate our AI detectors (GPTZero 5.0b, GPTZero 4.11b (aka 4.1o)), Pangram 4 and other open-source detectors on 11 popularly cited public AI detection benchmarks published by independent research groups. 

\textbf{AI in Peer Review}~\cite{saha2026peerreview} AI in Peer Review evaluates AI
detectors across the ways researchers may use AI to write or polish scientific peer
reviews, or use a humanizer to rewrite AI-generated reviews. The released benchmark
contains 50,155 reviews and covers both straightforward AI generation and more
challenging human--AI collaboration writing.

\textbf{Sem-Detect}~\cite{pmlr-v306-duarte26a} Sem-Detect is a benchmark of human and
AI-written (Gemini 2.5 Flash and Pro, DeepSeek V3.1 and Qwen3-235B-A22B) peer reviews
for ICLR and NeurIPS papers. It includes human reviews from OpenReview, fully
AI-generated reviews and human reviews refined by AI. AI reviews were written based on
the paper and the original human reviewer's score; refined reviews only polish the
original human reviewer's wording.

\textbf{DetectRL}~\cite{NEURIPS2024_b61bdf7e} DetectRL is a benchmark of human and
AI-written English text covering arXiv abstracts, BBC news, Reddit Writing Prompts
stories and Yelp reviews. The AI texts were generated by GPT-3.5-turbo, PaLM-2-bison,
Claude-instant and Llama-2-70b. It includes paraphrasing, AI polishing,
back-translation, small text edits and mixtures of human and AI sentences. It tests how
detectors perform across different sources, models, text lengths and edits to both human
and AI writing.

\textbf{UChicago}~\cite{jabarian2025artificial} UChicago DetectionAI evaluates leading
commercial and open-source AI detectors across diverse writing genres, text lengths, and
LLMs. The benchmark includes human writing, AI writing, and humanized AI writing across
blogs, consumer and restaurant reviews, news, novels, and r\'{e}sum\'{e}s.

\textbf{GEDE}~\cite{gehring2025gede} GEDE is a benchmark of human and AI-written English
student essays covering PERSUADE, BAWE (British Academic Written English) and AAE
(Argument Annotated Essays). It includes light edits and full rewrites of student
essays, essays generated from assignment prompts or summaries, and AI-generated essays
rewritten to bypass detection. Evaluation was performed on GEDESub, a 569-essay subset
covering human, fully AI-generated, AI-improved human and humanized AI essays.

\textbf{Epoch AI}~\cite{lee2026epochdetectors} Epoch AI built a benchmark of scientific
papers, blogs, and short fictional works using frontier LLMs including Claude-Opus-4.8,
GPT-5.5, and Gemini-3.1-Pro-Preview. Beyond direct AI generation from a generic prompt,
it includes a 5-shot style-imitation setting in which each LLM receives five real works
by the same author and is asked to imitate that author's writing style.

\textbf{GRiD}~\cite{Qazi_2024} GRiD is a benchmark of human and AI-written
(GPT-3.5-turbo) comments from the Reddit communities AskHistorians, AskScience and ELI5
(Explain Like I'm Five). The AI responses are capped at 100 tokens, making this a
benchmark of short, informal question-and-answer writing.

\textbf{Academic Integrity Abstracts}~\cite{karr2026academic} This work investigates a
failure mode of AI detection in academic integrity by revealing that detectors measure
statistical language patterns instead of authorship provenance. Its benchmark contains
published human research abstracts, Gemini-3-Flash rewrites of those abstracts, and
Gemini-3-Flash regenerations written from the source articles. The papers span
chemistry, computer science, political science, and theology. The source study includes
abstracts from 2013--2015 and 2023--2025. This evaluation uses only the 2013--2015 Human
cohort and its derived AI variants.

\textbf{MELD}~\cite{li2026meld} MELD (Multi-Task Equilibrated Learning Detector) is an
open-source model designed to detect AI-generated text even under adversarial attacks.
The benchmark released with this work uses RAID human source text from books, news,
scientific abstracts, recipes, Reddit, reviews, Wikipedia, and poetry. AI texts are
generated by GPT-5.4-Mini, Gemini-3-Flash, Claude-Haiku-4.5, and Qwen-3.6-Plus, then
transformed by six RAID-style attacks.

\textbf{Human Detector Russell}~\cite{russell2025human} This work demonstrates that
people with abundant LLM writing experience can identify AI-generated content and
compares human detection with automatic detection. The study releases 300 articles,
including human nonfiction writing, AI writing generated from paired human article
titles and subtitles, and humanized AI writing.

\textbf{VUB}~\cite{vanvlasselaer2026who} VUB Deep Research evaluates AI detectors on
master's thesis writing. The full study contains 160 papers spanning student thesis
writing, papers generated fully with OpenAI Deep Research, and two forms of human--AI
hybrid writing. Because of licensing restrictions, the human sources are not public, and
only 39 fully AI-generated papers are available.

\textbf{Results.} Table~\ref{tab:third-party} gives a breakdown of the results.
GPTZero 5.0b achieves state-of-the-art results on 9 of the 11 benchmarks, where it is
best or tied-best on every reported metric against Pangram 4 and the open-source
baselines: AI in Peer Review, Sem-Detect, DetectRL, GEDE, GRiD, Academic Integrity
Abstracts, MELD, Human Detector Russell and VUB. On the remaining two, UChicago and
Epoch AI, it trails on a single reported metric.

GPTZero 5.0b improves over its predecessor, GPTZero 4.11b, on most benchmarks and keeps
the false-positive rate near zero on every benchmark that reports one in
Table~\ref{tab:third-party}, with a single false positive on each of MELD (of 7,862),
Human Detector Russell (of 150) and Epoch AI (of 495), while maintaining a high AI
detection rate. At 1\% FPR, it detects every AI text in MELD across four generators and
six character-level attacks (220,136 texts) and every AI-generated review in
Sem-Detect. It also obtains the highest F1 in all eight DetectRL settings and on
average (97.34\%), generalizing across domains, generators and attacks.

The largest differences between detectors arise on documents that have been edited by
an LLM or passed through a humanizer. On AI in Peer Review, GPTZero 5.0b detects
98.20\% of humanized AI reviews, 96.96\% of AI-edited human reviews and 90.20\% of
humanized AI-edited reviews, compared with 78.50\%, 58.70\% and 71.70\% for the
strongest external detector. Such texts also account for the largest gains over
GPTZero 4.11b: 83.08\% vs.\ 63.44\% on short humanized UChicago texts, and 88.95\%
vs.\ 79.47\% on AI-improved human essays in GEDE. On GRiD, GPTZero 5.0b detects
99.93\% of AI answers at 1\% FPR and obtains the highest F1 at that threshold
(99.38\%), compared with 99.63\% and 98.31\% for the next-best detector. On Epoch AI,
Pangram 4 and Binoculars misclassify none of the 495 human texts, compared with one for
GPTZero 5.0b (0.20\%), while GPTZero 5.0b has the lowest false-negative rate (1.18\%).

The open-source baselines perform substantially worse overall, since they were
developed before most of the generators evaluated here. The strongest of them on
DetectRL, RADAR, averages 81.64\% F1, and Binoculars detects 43.75\% of standard
UChicago texts at 1\% FPR.


\subsection{Domain-Specific Performance}

To measure the detector's generalizability, we evaluate it on 18 human domain benchmarks covering academic, professional, creative and online writing, which include 2,933,419 human documents, between 19,106 and 200,000 per domain. The AI detection rate is measured on 465,045 AI texts from our evaluation sets, counting any non-human prediction, including mixed, as a detection.

Compared with its predecessor, GPTZero 4.11b, GPTZero 5.0b more than halves the overall false-positive rate and lowers it in nearly every domain (Table~\ref{tab:domain-fpr}), while also improving AI recall in most domains (Table~\ref{tab:domain-recall}).

\begin{table}[!htbp]
\centering
\small
\begin{adjustbox}{max width=\linewidth}
\begin{tabular}{lrrr}
\toprule
Domain & $n$ & 5.0b & 4.11b \\
\midrule
Admissions essay & 19,106 & \textbf{0.021} & 0.058 \\
Argumentative essay & 67,130 & \textbf{0.009} & 0.013 \\
Blog post & 200,000 & \textbf{0.021} & 0.038 \\
Descriptive essay & 200,000 & \textbf{0.026} & 0.050 \\
Email & 107,867 & \textbf{0.033} & 0.091 \\
Encyclopedia article & 200,000 & \textbf{0.067} & 0.168 \\
Expository essay & 200,000 & \textbf{0.039} & 0.077 \\
Financial disclosure & 200,000 & \textbf{0.008} & 0.035 \\
Lab report & 89,470 & \textbf{0.044} & 0.123 \\
Legal brief & 200,000 & \textbf{0.006} & 0.026 \\
Marketing copy & 200,000 & \textbf{0.018} & 0.042 \\
News article & 200,000 & \textbf{0.018} & 0.035 \\
Persuasive essay & 200,000 & \textbf{0.021} & 0.036 \\
Poem & 78,638 & \textbf{0.034} & 0.079 \\
Product review & 200,000 & 0.016 & \textbf{0.015} \\
Research paper & 200,000 & \textbf{0.017} & 0.037 \\
Service review & 171,208 & \textbf{0.016} & 0.032 \\
Short story & 200,000 & \textbf{0.013} & 0.037 \\
\bottomrule
\end{tabular}
\end{adjustbox}
\caption{False-positive rate (\%, $\downarrow$) of GPTZero 5.0b and 4.11b on the 18 human domain benchmarks, on identical samples: the share of human texts predicted as not human. Lower of the two models in bold.}
\label{tab:domain-fpr}
\end{table}

\begin{table}[!htbp]
\centering
\small
\begin{adjustbox}{max width=\linewidth}
\begin{tabular}{lrrr}
\toprule
Domain & $n$ & 5.0b & 4.11b \\
\midrule
Admissions essay & 859 & \textbf{99.53} & \textbf{99.53} \\
Argumentative essay & 1,350 & \textbf{99.63} & \textbf{99.63} \\
Blog post & 58,126 & \textbf{99.73} & 99.66 \\
Descriptive essay & 20,266 & \textbf{99.64} & 99.53 \\
Email & 4,745 & \textbf{98.23} & \textbf{98.23} \\
Encyclopedia article & 34,805 & \textbf{99.64} & 99.49 \\
Expository essay & 33,758 & \textbf{99.59} & 99.56 \\
Financial disclosure & 7,236 & \textbf{99.68} & 99.41 \\
Lab report & 4,389 & \textbf{99.82} & 99.64 \\
Legal brief & 27,482 & 98.92 & \textbf{99.02} \\
Marketing copy & 30,099 & \textbf{99.10} & 98.94 \\
News article & 85,222 & \textbf{99.55} & 99.36 \\
Persuasive essay & 14,284 & \textbf{99.70} & 99.68 \\
Poem & 4,240 & \textbf{98.63} & 98.54 \\
Product review & 20,067 & \textbf{99.75} & 99.65 \\
Research paper & 79,323 & \textbf{99.81} & 99.68 \\
Service review & 6,525 & \textbf{99.34} & 99.20 \\
Short story & 32,269 & \textbf{99.39} & 99.33 \\
\bottomrule
\end{tabular}
\end{adjustbox}
\caption{AI detection rate (\%, $\uparrow$) of GPTZero 5.0b and 4.11b by domain, on identical samples: the share of AI texts from our evaluation sets, grouped by domain, predicted as anything other than human. Higher of the two models in bold.}
\label{tab:domain-recall}
\end{table}

\subsection{Performance by Length and Generating Model}
\label{sec:length-llm}

In Table~\ref{tab:length}, we report performance of our AI detector grouped by document length (in words). The false-positive rate of GPTZero 5.0b falls from 0.116\% at 50--99 words to below 0.001\% from 500 words, which is consistently below GPTZero 4.11b across all length ranges.

\begin{table}[!htbp]
\centering
\small
\setlength{\tabcolsep}{3pt}
\begin{adjustbox}{max width=\linewidth}
\begin{tabular}{lrrrrrr}
\toprule
 & \multicolumn{3}{c}{FPR (\%) $\downarrow$} & \multicolumn{3}{c}{AI recall (\%) $\uparrow$} \\
\cmidrule(lr){2-4}\cmidrule(lr){5-7}
Words & $n$ & 5.0b & 4.11b & $n$ & 5.0b & 4.11b \\
\midrule
50--99 & 255,804 & \textbf{0.116} & 0.256 & 46,517 & \textbf{97.03} & 96.89 \\
100--199 & 482,359 & \textbf{0.045} & 0.095 & 51,405 & \textbf{99.17} & 98.97 \\
200--299 & 361,847 & \textbf{0.022} & 0.054 & 36,312 & \textbf{99.73} & 99.62 \\
300--499 & 488,329 & \textbf{0.006} & 0.015 & 53,927 & \textbf{99.86} & 99.81 \\
500--999 & 696,744 & \textbf{$<$0.001} & 0.001 & 98,838 & \textbf{99.92} & 99.86 \\
1000+ & 628,409 & \textbf{$<$0.001} & 0.001 & 96,740 & \textbf{99.91} & 99.81 \\
\bottomrule
\end{tabular}
\end{adjustbox}
\caption{Performance by document length. FPR on the human texts of the 18 domain benchmarks; AI recall (predicted exactly AI) on AI texts in prose domains. Texts of at least 50 words. Better of the two models in bold.}
\label{tab:length}
\end{table}

In Table~\ref{tab:llms} we also evaluate our model's AI recall across on documents generated from a diverse set of providers including OpenAI, Anthropic, Google, xAI, DeepSeek, Alibaba, Moonshot and NVIDIA. GPTZero 5.0b achieves higher detection rates than GPTZero 4.11b on nearly all 50 generating models, including the newest frontier models.

\begin{table*}[!tp]
\centering
\small
\setlength{\tabcolsep}{4pt}
\renewcommand{\arraystretch}{1.1}
\begin{adjustbox}{max width=\linewidth}
\begin{tabular}{llrrr@{\hskip 20pt}llrrr}
\toprule
Provider & Generating model & $n$ & 5.0b & 4.11b & Provider & Generating model & $n$ & 5.0b & 4.11b \\
\midrule
OpenAI & GPT-6.1 Sol$^{\star}$ & 4,822 & \textbf{99.61} & 98.90 & Anthropic & Claude Opus 5.5$^{\star}$ & 4,710 & \textbf{99.83} & 99.34 \\
 & GPT-6 Astra$^{\star}$ & 4,791 & \textbf{99.48} & 99.21 &  & Claude Fable 5.1$^{\star}$ & 4,922 & \textbf{99.88} & 99.59 \\
 & GPT-5.6 Terra & 13,544 & \textbf{99.68} & 99.50 &  & Claude Sonnet 5 & 3,988 & \textbf{99.87} & 99.75 \\
 & GPT-5.6 Sol & 13,626 & \textbf{99.72} & 99.66 &  & Claude Opus 5$^{\dagger}$ & 5,664 & \textbf{99.89} & 99.51 \\
 & GPT-5.6 Luna & 3,574 & \textbf{99.97} & 99.94 &  & Claude Fable 5$^{\dagger}$ & 5,680 & \textbf{99.77} & 99.37 \\
 & GPT-5.5 & 10,138 & \textbf{99.74} & 99.64 &  & Claude Sonnet 4.6 & 7,609 & \textbf{99.62} & 99.38 \\
 & GPT-5.4 mini & 7,834 & \textbf{100.00} & 99.99 &  & Claude Opus 4.6 & 8,857 & \textbf{99.28} & 98.78 \\
 & GPT-5.4 & 6,315 & \textbf{99.52} & 99.38 &  & Claude Sonnet 4.5 & 5,950 & \textbf{99.18} & 98.67 \\
 & GPT-5.2 & 152,528 & \textbf{98.83} & \textbf{98.83} &  & Claude Opus 4.5 & 129,751 & \textbf{99.73} & 99.52 \\
 & GPT-5 nano & 16,984 & \textbf{99.42} & 99.39 &  & Claude Sonnet 4 & 7,161 & \textbf{99.62} & 99.47 \\
 & GPT-5 mini & 36,992 & \textbf{99.52} & 99.51 &  & Claude 3.7 Sonnet & 6,911 & \textbf{98.97} & 98.58 \\
 & GPT-5 & 67,336 & \textbf{99.76} & 99.73 &  & Claude Haiku (unversioned) & 7,778 & \textbf{100.00} & 99.99 \\\cline{6-10}
 & GPT-4.1 mini & 23,122 & \textbf{99.77} & 99.59 & Google & Gemini 4 Argon$^{\star}$ & 5,285 & \textbf{99.30} & 98.28 \\
 & GPT-4.1 & 22,951 & \textbf{99.61} & 99.29 &  & Gemini 3.6 Flash & 18,420 & 99.79 & \textbf{99.83} \\
 & GPT-4o mini & 25,538 & \textbf{99.85} & 99.82 &  & Gemini 3.1 Pro & 8,334 & \textbf{99.15} & 98.84 \\
 & GPT-4o & 37,701 & \textbf{99.87} & 99.79 &  & Gemini 3.1 Flash-Lite & 8,939 & \textbf{99.82} & 99.60 \\
 & GPT-3.5 Turbo & 1,368 & \textbf{99.78} & 97.22 &  & Gemini 3 Pro & 127,191 & \textbf{99.42} & 99.34 \\
 & o3-mini & 4,833 & \textbf{99.83} & 99.61 &  & Gemini 3 Flash & 7,821 & \textbf{100.00} & \textbf{100.00} \\
 & o3 & 3,840 & \textbf{99.69} & 99.43 &  & Gemini 2.5 Pro & 8,590 & \textbf{99.16} & 98.82 \\\cline{1-5}
xAI & Grok 4.7$^{\star}$ & 5,154 & \textbf{99.83} & 99.67 &  & Gemini 2.5 Flash-Lite & 5,818 & \textbf{99.40} & 98.92 \\
 & Grok 4.5$^{\dagger}$ & 5,709 & \textbf{99.77} & 99.44 &  & Gemini 2.5 Flash & 9,720 & \textbf{99.57} & 99.39 \\
 & Grok 4 Fast & 46,695 & \textbf{99.74} & 99.63 &  & Gemini 2.0 Flash & 6,910 & \textbf{98.76} & 98.10 \\\cline{1-5}\cline{6-10}
DeepSeek & DeepSeek V4.1 Flash 1017$^{\star}$ & 4,784 & \textbf{99.77} & 99.41 & Moonshot & Kimi K3$^{\star}$ & 5,178 & \textbf{99.92} & 99.67 \\\cline{6-10}
 & DeepSeek V3.2 & 9,954 & \textbf{98.51} & 98.12 & NVIDIA & Nemotron 3.5 Super VL$^{\star}$ & 5,002 & \textbf{99.90} & 99.76 \\\cline{1-5}
Alibaba & Qwen3.8 27B$^{\star}$ & 4,611 & \textbf{99.93} & 99.83 &  & & & &  \\
 & Qwen 3.6 Plus & 7,810 & \textbf{100.00} & \textbf{100.00} &  & & & &  \\
\bottomrule
\end{tabular}
\end{adjustbox}
\caption{AI detection rate (\%, $\uparrow$) of GPTZero 5.0b and 4.11b by generating model: the share of AI texts predicted as anything other than human, pooled across our evaluation sets, on identical samples. Better of the two models in bold. $^{\star}$Newest models, evaluated on responses to WildChat prompts. $^{\dagger}$Evaluated on a separate evaluation set.}
\label{tab:llms}
\end{table*}

\subsection{Fairness and False Positives on Learner Corpora}
\label{sec:fairness}
We evaluate our detector on six corpora of writing by L1 and L2 English writers and one
corpus of school essays annotated with student demographics. L1 writers use English as
their first language, while L2 writers use English as an additional language.
\paragraph{ASAP 2.0~\cite{crossley2025asap}} contains source-based argumentative essays
written by US students in grades 6, 8, 9 and 10 for state writing assessments, and
records English Language Learner (ELL) status alongside other student demographics. We
evaluate 4,813 essays of at least 300 words: 4,072 by L1 writers, 573 by L2 writers and
168 with unknown ELL status.
\paragraph{PELIC~\cite{naismith2022pelic}} is a 4.2-million-word corpus of written
coursework by 1,177 students in the University of Pittsburgh's Intensive English
Program, collected from 2006 to 2012. The most common first languages are Arabic,
Chinese, Japanese, Korean, Spanish and Turkish, and proficiency ranges from about CEFR A2
to B2/C1. We evaluate 4,269 L2 English texts of at least 300 words; the length threshold
removes the corpus's short exercises, such as word lists, vocabulary logs and grammar
drills.
\paragraph{CELL~\cite{leung2022cell}} is a corpus of academic essay assignments in
Chinese and English by university undergraduates, mostly from Hong Kong, most of whom
speak Cantonese as their first language. We evaluate 2,827 L2 English texts of at least
300 words, written from 2018 to 2020.
\paragraph{ELLIPSE~\cite{crossley2023ellipse}} is a corpus of 6,482 essays by English
language learners in grades 8 to 12, written for US statewide writing assessments in the
2018--19 and 2019--20 school years. We evaluate 5,107 essays of at least 300 words.
ELLIPSE and PERSUADE 2.0 come from the same essay collection, and over 300 ELLIPSE essays
are also in the PERSUADE 2.0 test split.
\paragraph{ICNALE~\cite{ishikawa2023icnale}} Written Essays (version 2.6) holds 5,600
essays by 2,600 college students in ten Asian countries and regions and 200 English
native speakers, collected from 2007 to 2013. Each writer wrote two 200--300-word opinion
essays, on part-time jobs for college students and on banning smoking in restaurants. We
evaluate 710 essays of at least 300 words: 25 by L1 writers and 685 by L2 writers.
\paragraph{ChatGPT-Detector-Bias~\cite{liang2023gpt}} accompanies the study that found
seven widely used AI detectors misclassified, on average, more than half of a set of TOEFL essays by
non-native English writers as AI-generated. It contains 91 TOEFL essays by L2 writers from
a Chinese educational forum, 88 US 8th-grade essays by L1 writers from the Hewlett
Foundation's ASAP dataset, and 70 college admission essays by L1 writers. We evaluate all
249 essays.
\paragraph{PERSUADE 2.0~\cite{crossley2024persuade}} is a corpus of over 25,000
argumentative essays by US students in grades 6 to 12, written for state and national
writing assessments from 2010 to 2020 in response to 15 prompts, each paired with the
student's demographics. We evaluate the release's test split of 10,402 essays and report
the false-positive rate for 23 student groups, defined by disability status, economic
disadvantage, English language learner status, gender, grade level and race or
ethnicity.

\begin{table}[!htbp]
\centering
\small
\setlength{\tabcolsep}{4pt}
\begin{adjustbox}{max width=\linewidth}
\begin{tabular}{llrrr}
\toprule
Corpus & Writers & Texts & 5.0b & 4.11b \\
\midrule
ASAP 2.0 & L1 & 4,072 & \textbf{0.00} & \textbf{0.00} \\
 & L2 & 573 & \textbf{0.00} & \textbf{0.00} \\
 & Unknown & 168 & \textbf{0.00} & \textbf{0.00} \\
\addlinespace
ICNALE & L1 & 25 & \textbf{0.00} & \textbf{0.00} \\
 & L2 & 685 & \textbf{0.00} & 0.15 \\
\addlinespace
PELIC & L2 & 4,269 & \textbf{0.00} & 0.05 \\
ELLIPSE & L2 & 5,107 & \textbf{0.00} & 0.04 \\
CELL & L2 & 2,827 & \textbf{0.00} & \textbf{0.00} \\
\bottomrule
\end{tabular}
\end{adjustbox}
\caption{False-positive rate (\%, $\downarrow$) of GPTZero 5.0b and 4.11b by learner corpus
on human-written texts of at least 300 words, on identical samples. L1: English as a first
language; L2: English as an additional language; Unknown: not recorded. Lower or equal of
the two models in bold.}
\label{tab:fairness-learner}
\end{table}

\begin{table}[!htbp]
\centering
\small
\setlength{\tabcolsep}{4pt}
\begin{adjustbox}{max width=\linewidth}
\begin{tabular}{lrrrrr}
\toprule
Writers & Texts & 5.0b & 4.11b & Pangram 4$^{\S}$ & Turnitin$^{\S}$ \\
\midrule
L2 English & 5,527 & \textbf{0.00} & 0.05 & 0.02 & 1.40 \\
L1 English & 4,097 & \textbf{0.00} & \textbf{0.00} & \textbf{0.00} & 1.30 \\
\bottomrule
\end{tabular}
\end{adjustbox}
\caption{False-positive rate (\%, $\downarrow$) on human-written texts of at least 300 words
by L1 and L2 English writers, following Turnitin's evaluation
protocol~\cite{turnitin2023ell}. GPTZero results pool ASAP 2.0, ICNALE and PELIC, with
GPTZero 5.0b and 4.11b evaluated on identical samples. $^{\S}$Aggregate rates as published
by Pangram~\cite{pangram2026esl} and Turnitin~\cite{turnitin2023ell} on their own samples.
Lower value per row in bold.}
\label{tab:fairness-learner-pooled}
\end{table}

\begin{table}[!htbp]
\centering
\small
\setlength{\tabcolsep}{4pt}
\begin{adjustbox}{max width=\linewidth}
\begin{tabular}{lrrr}
\toprule
Group & Essays & 5.0b & 4.11b \\
\midrule
\multicolumn{4}{l}{\textit{Identified as having a disability}} \\
\quad Yes & 1,172 & \textbf{0.00 (0)} & \textbf{0.00 (0)} \\
\quad No & 7,559 & \textbf{0.00 (0)} & 0.03 (2) \\
\quad Unknown & 1,671 & \textbf{0.00 (0)} & \textbf{0.00 (0)} \\
\addlinespace
\multicolumn{4}{l}{\textit{Economically disadvantaged}} \\
\quad Yes & 4,252 & \textbf{0.00 (0)} & 0.02 (1) \\
\quad No & 4,447 & \textbf{0.00 (0)} & 0.02 (1) \\
\quad Unknown & 1,703 & \textbf{0.00 (0)} & \textbf{0.00 (0)} \\
\addlinespace
\multicolumn{4}{l}{\textit{English language learner}} \\
\quad Yes & 914 & \textbf{0.00 (0)} & \textbf{0.00 (0)} \\
\quad No & 8,984 & \textbf{0.00 (0)} & 0.02 (2) \\
\quad Unknown & 504 & \textbf{0.00 (0)} & \textbf{0.00 (0)} \\
\addlinespace
\multicolumn{4}{l}{\textit{Gender}} \\
\quad Female & 5,260 & \textbf{0.00 (0)} & 0.04 (2) \\
\quad Male & 5,142 & \textbf{0.00 (0)} & \textbf{0.00 (0)} \\
\addlinespace
\multicolumn{4}{l}{\textit{Grade level}} \\
\quad 6 & 684 & \textbf{0.00 (0)} & \textbf{0.00 (0)} \\
\quad 8 & 4,015 & \textbf{0.00 (0)} & 0.05 (2) \\
\quad 9 & 235 & \textbf{0.00 (0)} & \textbf{0.00 (0)} \\
\quad 10 & 3,620 & \textbf{0.00 (0)} & \textbf{0.00 (0)} \\
\quad 11 & 1,220 & \textbf{0.00 (0)} & \textbf{0.00 (0)} \\
\quad 12 & 161 & \textbf{0.00 (0)} & \textbf{0.00 (0)} \\
\quad Unknown & 467 & \textbf{0.00 (0)} & \textbf{0.00 (0)} \\
\addlinespace
\multicolumn{4}{l}{\textit{Race/ethnicity}} \\
\quad Asian/Pacific Islander & 671 & \textbf{0.00 (0)} & \textbf{0.00 (0)} \\
\quad Black/African American & 1,984 & \textbf{0.00 (0)} & \textbf{0.00 (0)} \\
\quad Hispanic/Latino & 2,691 & \textbf{0.00 (0)} & \textbf{0.00 (0)} \\
\quad Two or more races/Other & 424 & \textbf{0.00 (0)} & \textbf{0.00 (0)} \\
\quad White & 4,559 & \textbf{0.00 (0)} & 0.04 (2) \\
\bottomrule
\end{tabular}
\end{adjustbox}
\caption{False-positive rate (\%, $\downarrow$; false positives in parentheses) of GPTZero 5.0b and 4.11b on 10,402 PERSUADE 2.0 student essays, by student group, on identical samples. GPTZero 5.0b makes no false positive in any group. Lower or equal of the two models in bold. Groups below 100 essays are omitted, missing attributes are marked Unknown.}
\label{tab:fairness-persuade}
\end{table}

\begin{table}[!htbp]
\centering
\small
\setlength{\tabcolsep}{4pt}
\begin{adjustbox}{max width=\linewidth}
\begin{tabular}{llrrr}
\toprule
Essays & Writers & Texts & 5.0b & 4.11b \\
\midrule
TOEFL & L2 & 91 & \textbf{0.00 (0)} & 1.10 (1) \\
US 8th grade (Hewlett) & L1 & 88 & \textbf{0.00 (0)} & \textbf{0.00 (0)} \\
College admissions & L1 & 70 & \textbf{0.00 (0)} & \textbf{0.00 (0)} \\
\bottomrule
\end{tabular}
\end{adjustbox}
\caption{False-positive rate (\%, $\downarrow$; false positives in parentheses) of GPTZero
5.0b and 4.11b on all essays of the ChatGPT-Detector-Bias benchmark~\cite{liang2023gpt},
on identical samples. Lower or equal of the two models in bold.}
\label{tab:fairness-liang}
\end{table}

\paragraph{Results.} Following Turnitin's public evaluation
protocol~\cite{turnitin2023ell}, we evaluate human-written texts of at least 300 words and
report false-positive rates for L1 and L2 English writers. Pooled over ASAP 2.0, ICNALE and
PELIC (Table~\ref{tab:fairness-learner-pooled}), GPTZero 5.0b's false-positive rate is
0.00\% for both L2 and L1 writers, compared with 0.05\% and 0.00\% for GPTZero 4.11b and
with the 0.02\% and 0.00\% published for Pangram~4~\cite{pangram2026esl} and the 1.40\%
and 1.30\% published for Turnitin~\cite{turnitin2023ell}. GPTZero 5.0b makes no false
positive on any individual learner corpus either (Table~\ref{tab:fairness-learner}),
including CELL and ELLIPSE. On ChatGPT-Detector-Bias (Table~\ref{tab:fairness-liang}), it
makes no false positive on any of the 249 essays, including all 91 TOEFL essays, of which
GPTZero 4.11b misclassifies one. On PERSUADE 2.0 (Table~\ref{tab:fairness-persuade}),
GPTZero 5.0b makes no false positive in any of the 23 student groups. On PERSUADE 2.0 (Table~\ref{tab:fairness-persuade}), GPTZero
5.0b makes no false positive in any of the 23 student groups.

\subsection{AI Bypasser Detection}
\label{section:ai_bypassers}

AI bypassers are at the frontier of AI detection, offering sophisticated users an opportunity to humanize a text without manual intervention. This is a particularly challenging setting.
We use a dataset of 1000 AI texts that have been modified by a variety of paraphrasing techniques and 11 dedicated bypassers services (see Appendix Section \ref{appendix:paraphraser_services} for details and Table~\ref{table:paraphraser_services}) to quantify accuracy in this setting \cite{krishna2023paraphrasing}. This dataset covers several domains including academic writing, scientific text, essays, social media, and creative writing. The original AI text was generated by GPT-5, GPT4o, and GPT4.1. Counting a text as detected when it is classified as AI-generated or AI-paraphrased, GPTZero 5.0b detects 95.4\% of the bypassed texts, compared with 89.9\% for Pangram 4. When mixed predictions are also counted as detections, GPTZero 5.0b reaches 95.5\% and Pangram 4 reaches 94.6\%. These results demonstrate the effectiveness of our multi-tiered red teaming approach.

\subsection{Polished Text Detection}
\begin{figure}[h!]
\centering
\includegraphics[scale=0.2]{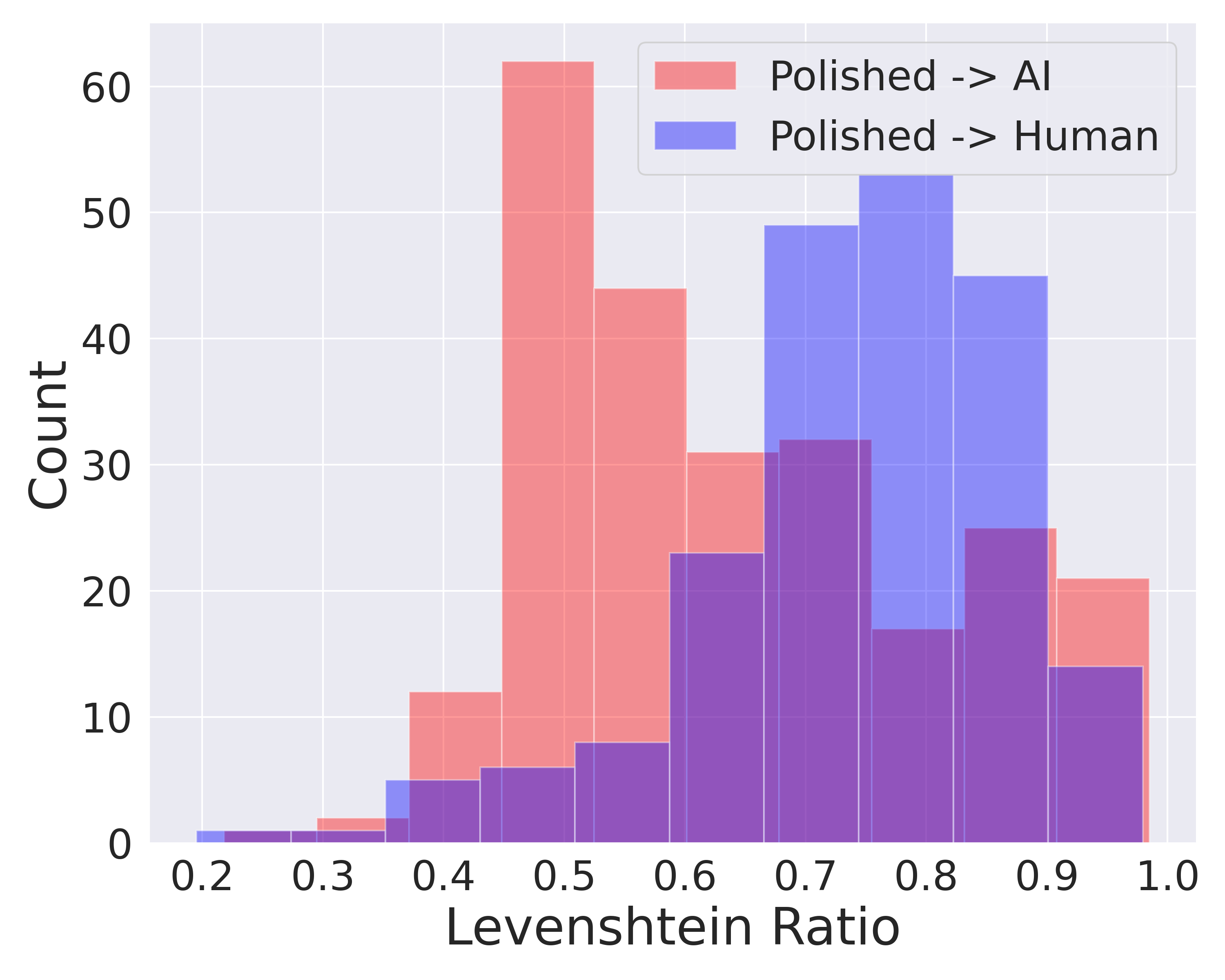}
\caption{Levensthein ratio for polished sample misclassifications.}
\label{figure:levensthein_ratio}
\end{figure}

The definition of a Polished text presented in Section \ref{sec:polished_method} is unique to \gptzero{} since competitors which offer such a service may use other thresholds and prompts. As a result, we measure \gptzero{}'s ability to detect polished texts on a larger dataset and exclude competitors from this analysis since it would be confounded by a difference in definition. We use a dataset of 4,631 polished texts across several human origin datasets. Of these, we correctly classify 4,175 as polished, and misclassify 247 as AI, 205 as human, and 4 as mixed. Figure \ref{figure:levensthein_ratio} shows the Levenshtein ratio relative to the original human text for the misclassifications where samples misclassified as AI are less similar to the human text on average, compared to samples misclassified as human.

\subsection{Mixed Class Ablation Study}
\begin{figure}[h!]
\centering
\includegraphics[scale=0.25]{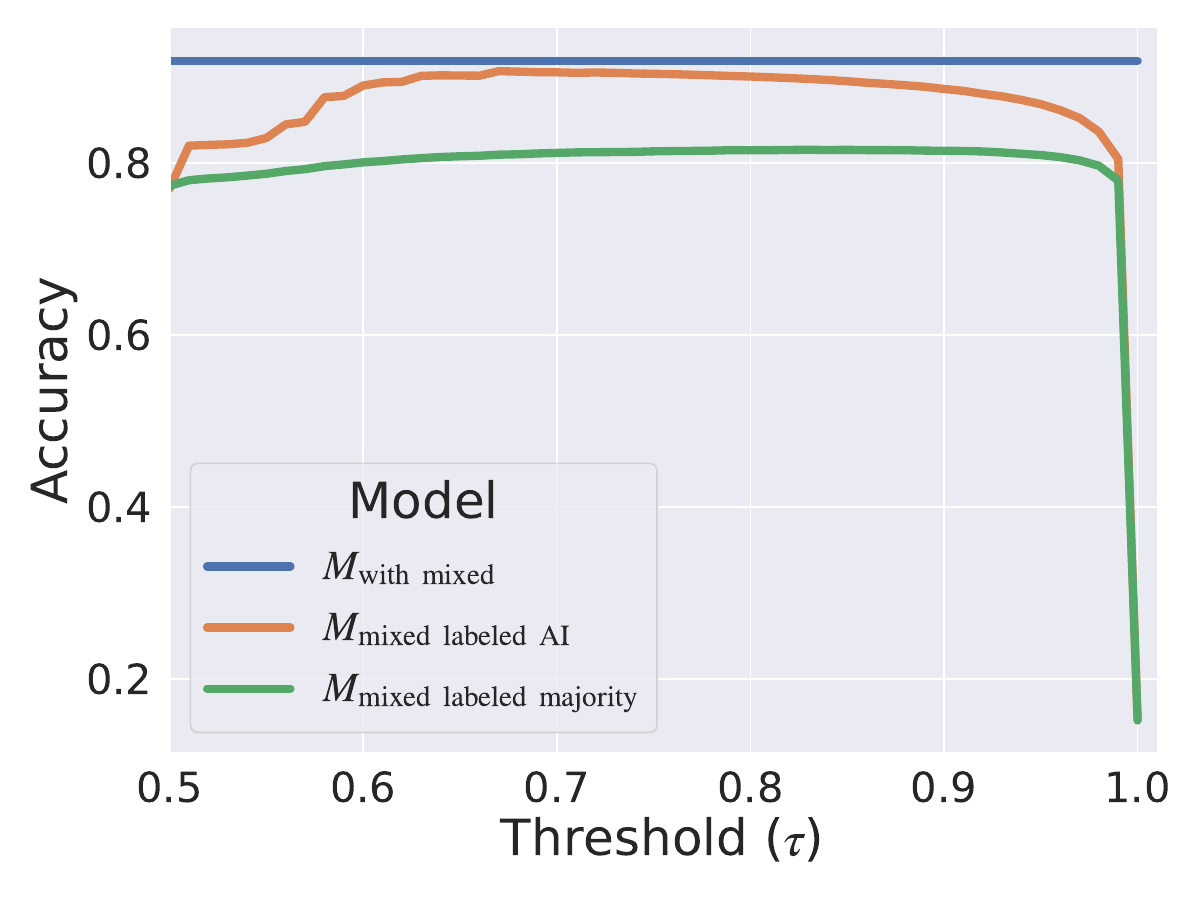}
\caption{Accuracy of mixed class ablation models on classifying documents evaluated on internal dataset of 40k examples (details in Appendix).
}
\label{figure:mixed_ablation}
\end{figure}

We investigate the role that training on 3 classes has on model performance. We consider 3 ablations where during training samples containing both human and AI text are labeled as: AI (in the \mixedasai{} setting), the majority sentence proportion label (in the \mixedasmajority), and mixed (in the \mixednormal{} category - Ours). All 3 of these approaches achieve similar AUC ($\sim$96\%) when evaluated on purely Human/AI documents. However, forcing a ternary classification problem to be binary requires using sentence-level predictions to classify if a document contains a mix of human and AI text. The following logic is used: given the average sentence-level AI probability $p_{\mathrm{avg}} = \frac{1}{|S|} \sum_{i=1}^{|S|} p_{ai}(s_{i})$, the predicted class is
$f(p_{\mathrm{avg}}) = \text{Human if } p_{\mathrm{avg}} < 1-\tau; \text{Mixed if } 1-\tau \le p_{\mathrm{avg}} \le \tau; \text{AI if } p_{\mathrm{avg}} > \tau.$ for $\tau \in [0.5, 1]$

Figure \ref{figure:mixed_ablation} shows that our mixed class approach is superior when considering performance on all document types (Human, AI, Mixed), with the added benefit of not having to perform a hyperparameter search for $\tau$ since we use $\argmax$ over our multiclass probabilites to classify.

\section{Conclusion}
With a focus on meticulous data gathering, generation, and augmentation efforts, as well innovations pertaining to model training, architecture, and inference, we have achieved a state of the art commercial AI text detector. Our approach has been comprehensively evaluated on various domains, LLMs, generation settings, prompts, and even possible paraphrasing, demonstrating broad applicability. The transparency provided by \gptzero{} enables users to make informed decisions about trusting content and evaluating the effort expended to generate it. 

\section{Limitations}
Our work shares some limitations with other research on AI detection. For example, a challenging and representative evaluation dataset capable of revealing model shortfalls is elusive. 
We find that in-distribution performance metrics are overly optimistic and unable to distinguish between models with significantly different OOD performance. 
Additionally, there is a lack of standardization regarding which datasets are used to evaluate models both in the research literature, and by commercial providers. 
This introduces the risk of cherry-picking, and makes it challenging to determine if newly introduced methods are better than existing ones. 
We believe a public benchmark with regular updates incorporating new LLM versions would be highly beneficial to the AI-generated text detection community as a whole \cite{Kiela2021-tx}. 

Developing an accurate and scalable deep learning model for AI-generated text detection requires meticulous data engineering efforts. 
This data engineering can be resource intensive and still result in blind spots, such as poor model generalization to lower quality LLMs like GPT-2, OPT, etc. 
Generalization to new models is of particular importance, as the distribution of user text changes over time with the release of newer and more capable LLMs. 
Being robust to users trying to bypass our detector also remains a priority, especially as some users incorporate paraphrasers and adversarial attacks as part of their LLM-based writing process. 
While techniques such as adversarial training improve robustness \cite{Tramer2019-gk}, it is known to trade-off performance on in-distribution examples \cite{Raghunathan2019-jw,Hu2023-cg}. 

Furthermore, our Advanced Scan method cannot entirely capture human text editing patterns. As such, it is possible that the most important sentences it identifies are suboptimal. However, the experiments showing the effect of removing important sentences do support that the presence of these sentences are at least correlated with detector predictions, even if they are not necessarily causal.





\bibliography{acl_latex}

@BOOK{Randall2001-dw,
  title     = "Pragmatic Plagiarism: Authorship, Profit, and Power",
  author    = "Randall, Marilyn",
  publisher = "University of Toronto Press",
  month     =  jan,
  year      =  2001,
  language  = "en"
}

@ARTICLE{Padillah2024-js,
  title    = "Ghostwriting: a reflection of academic dishonesty in the
              artificial intelligence era",
  author   = "Padillah, Raup",
  journal  = "J. Public Health",
  volume   =  46,
  number   =  1,
  pages    = "e193--e194",
  month    =  feb,
  year     =  2024,
  language = "en"
}

@misc{penedo2024finewebdatasetsdecantingweb,
      title={The FineWeb Datasets: Decanting the Web for the Finest Text Data at Scale},
      author={Guilherme Penedo and Hynek Kydlíček and Loubna Ben allal and Anton Lozhkov and Margaret Mitchell and Colin Raffel and Leandro Von Werra and Thomas Wolf},
      year={2024},
      eprint={2406.17557},
      archivePrefix={arXiv},
      primaryClass={cs.CL},
      url={https://arxiv.org/abs/2406.17557},
}

@ARTICLE{Yang2024-pz,
  title     = "Anatomy of an {AI-powered} malicious social botnet",
  author    = "Yang, Kaicheng and Menczer, Filippo",
  journal   = "J. Quant. Descr. Digit. Media",
  publisher = "Journal of Quantitative Description: Digital Media",
  volume    =  4,
  month     =  may,
  year      =  2024
}

@ARTICLE{Liang2024-iu,
  title         = "Mapping the Increasing Use of {LLMs} in Scientific Papers",
  author        = "Liang, Weixin and Zhang, Yaohui and Wu, Zhengxuan and Lepp,
                   Haley and Ji, Wenlong and Zhao, Xuandong and Cao, Hancheng
                   and Liu, Sheng and He, Siyu and Huang, Zhi and Yang, Diyi
                   and Potts, Christopher and Manning, Christopher D and Zou,
                   James Y",
  month         =  apr,
  year          =  2024,
  archivePrefix = "arXiv",
  primaryClass  = "cs.CL",
  eprint        = "2404.01268"
}

@ARTICLE{Lai2024-eh,
  title         = "Adaptive Ensembles of {Fine-Tuned} Transformers for
                   {LLM-Generated} Text Detection",
  author        = "Lai, Zhixin and Zhang, Xuesheng and Chen, Suiyao",
  month         =  mar,
  year          =  2024,
  archivePrefix = "arXiv",
  primaryClass  = "cs.LG",
  eprint        = "2403.13335"
}

@ARTICLE{Tulchinskii2023-mb,
  title         = "Intrinsic Dimension Estimation for Robust Detection of
                   {AI-Generated} Texts",
  author        = "Tulchinskii, Eduard and Kuznetsov, Kristian and Kushnareva,
                   Laida and Cherniavskii, Daniil and Barannikov, Serguei and
                   Piontkovskaya, Irina and Nikolenko, Sergey and Burnaev,
                   Evgeny",
  month         =  jun,
  year          =  2023,
  archivePrefix = "arXiv",
  primaryClass  = "cs.CL",
  eprint        = "2306.04723"
}

@ARTICLE{Guo2017-wo,
  title         = "On Calibration of Modern Neural Networks",
  author        = "Guo, Chuan and Pleiss, Geoff and Sun, Yu and Weinberger,
                   Kilian Q",
  month         =  jun,
  year          =  2017,
  archivePrefix = "arXiv",
  primaryClass  = "cs.LG",
  eprint        = "1706.04599"
}

@UNPUBLISHED{Bao2023-ay,
  title       = "{Fast-DetectGPT}: Efficient {Zero-Shot} Detection of
                 {Machine-Generated} Text via Conditional Probability Curvature",
  author      = "Bao, Guangsheng and Zhao, Yanbin and Teng, Zhiyang and Yang,
                 Linyi and Zhang, Yue",
  institution = "arXiv [cs.CL]",
  month       =  oct,
  year        =  2023,
  language    = "en"
}

@ARTICLE{Hans2024-qb,
  title     = "Spotting {LLMs} with Binoculars: Zero-shot detection of
               machine-generated text",
  author    = "Hans, Abhimanyu and Schwarzschild, Avi and Cherepanova, Valeriia
               and Kazemi, Hamid and Saha, Aniruddha and Goldblum, Micah and
               Geiping, Jonas and Goldstein, Tom",
  journal   = "arXiv.org",
  publisher = "arXiv",
  year      =  2024,
  language  = "en"
}

@article{maung24,
  author  = "Maung, Barani Maung and McBride, Keegan and Lucas, Jason S. and Tabar, Maryam and Lee, Dongwon",
  title   = "Generative AI Disproportionately Harms Long Tail Users",
  journal = "IEEE Computer",
  year    = 2024,
  volume  = "57",
  number  = "11",
}

@ARTICLE{Su2023-de,
  title         = "{DetectLLM}: Leveraging Log Rank Information for {Zero-Shot}
                   Detection of {Machine-Generated} Text",
  author        = "Su, Jinyan and Zhuo, Terry Yue and Wang, Di and Nakov,
                   Preslav",
  month         =  may,
  year          =  2023,
  archivePrefix = "arXiv",
  primaryClass  = "cs.CL",
  eprint        = "2306.05540"
}

@ARTICLE{Mitchell2023-qb,
  title         = "{DetectGPT}: {Zero-Shot} {Machine-Generated} Text Detection
                   using Probability Curvature",
  author        = "Mitchell, Eric and Lee, Yoonho and Khazatsky, Alexander and
                   Manning, Christopher D and Finn, Chelsea",
  month         =  jan,
  year          =  2023,
  archivePrefix = "arXiv",
  primaryClass  = "cs.CL",
  eprint        = "2301.11305"
}

@ARTICLE{Raghunathan2019-jw,
  title         = "Adversarial Training Can Hurt Generalization",
  author        = "Raghunathan, Aditi and Xie, Sang Michael and Yang, Fanny and
                   Duchi, John C and Liang, Percy",
  month         =  jun,
  year          =  2019,
  archivePrefix = "arXiv",
  primaryClass  = "cs.LG",
  eprint        = "1906.06032"
}

@ARTICLE{Tramer2019-gk,
  title    = "Adversarial Training and Robustness for Multiple Perturbations",
  author   = "Tram{\`e}r, Florian and Boneh, D",
  journal  = "Adv. Neural Inf. Process. Syst.",
  pages    = "5858--5868",
  month    =  apr,
  year     =  2019
}

@inproceedings{Venkatraman2023-pe,
    title = "{GPT}-who: An Information Density-based Machine-Generated Text Detector",
    author = "Venkatraman, Saranya  and
      Uchendu, Adaku  and
      Lee, Dongwon",
    editor = "Duh, Kevin  and
      Gomez, Helena  and
      Bethard, Steven",
    booktitle = "Findings of the Association for Computational Linguistics: NAACL 2024",
    month = jun,
    year = "2024",
    address = "Mexico City, Mexico",
    pages = "103--115",
}

@ARTICLE{Islam2023-uj,
  title         = "Distinguishing Human Generated Text From {ChatGPT} Generated
                   Text Using Machine Learning",
  author        = "Islam, Niful and Sutradhar, Debopom and Noor, Humaira and
                   Raya, Jarin Tasnim and Maisha, Monowara Tabassum and Farid,
                   Dewan Md",
  month         =  may,
  year          =  2023,
  archivePrefix = "arXiv",
  primaryClass  = "cs.CL",
  eprint        = "2306.01761"
}

@ARTICLE{Hearst1998-ty,
  title     = "Support vector machines",
  author    = "Hearst, M A and Dumais, S T and Osuna, E and Platt, J and
               Scholkopf, B",
  journal   = "IEEE Intelligent Systems and their Applications",
  publisher = "IEEE",
  volume    =  13,
  number    =  4,
  pages     = "18--28",
  year      =  1998
}

@ARTICLE{Hochreiter1997-gp,
  title    = "Long short-term memory",
  author   = "Hochreiter, S and Schmidhuber, J",
  journal  = "Neural Comput.",
  volume   =  9,
  number   =  8,
  pages    = "1735--1780",
  month    =  nov,
  year     =  1997,
  language = "en"
}

@ARTICLE{Kumarage2023-db,
  title         = "Stylometric Detection of {AI-Generated} Text in Twitter
                   Timelines",
  author        = "Kumarage, Tharindu and Garland, Joshua and Bhattacharjee,
                   Amrita and Trapeznikov, Kirill and Ruston, Scott and Liu,
                   Huan",
  month         =  mar,
  year          =  2023,
  archivePrefix = "arXiv",
  primaryClass  = "cs.CL",
  eprint        = "2303.03697"
}

@ARTICLE{Hu2023-cg,
  title         = "{RADAR}: Robust {AI-text} detection via adversarial learning",
  author        = "Hu, Xiaomeng and Chen, Pin-Yu and Ho, Tsung-Yi",
  month         =  jul,
  year          =  2023,
  copyright     = "http://arxiv.org/licenses/nonexclusive-distrib/1.0/",
  archivePrefix = "arXiv",
  primaryClass  = "cs.CL",
  eprint        = "2307.03838"
}

@ARTICLE{Bhattacharjee2023-yv,
  title         = "{ConDA}: Contrastive Domain Adaptation for {AI-generated}
                   Text Detection",
  author        = "Bhattacharjee, Amrita and Kumarage, Tharindu and Moraffah,
                   Raha and Liu, Huan",
  month         =  sep,
  year          =  2023,
  archivePrefix = "arXiv",
  primaryClass  = "cs.CL",
  eprint        = "2309.03992"
}

@ARTICLE{Tian2023-ui,
  title         = "Multiscale {Positive-Unlabeled} Detection of {AI-Generated}
                   Texts",
  author        = "Tian, Yuchuan and Chen, Hanting and Wang, Xutao and Bai,
                   Zheyuan and Zhang, Qinghua and Li, Ruifeng and Xu, Chao and
                   Wang, Yunhe",
  month         =  may,
  year          =  2023,
  archivePrefix = "arXiv",
  primaryClass  = "cs.CL",
  eprint        = "2305.18149"
}

@ARTICLE{Dugan2024-qu,
  title         = "{RAID}: A Shared Benchmark for Robust Evaluation of
                   {Machine-Generated} Text Detectors",
  author        = "Dugan, Liam and Hwang, Alyssa and Trhlik, Filip and Ludan,
                   Josh Magnus and Zhu, Andrew and Xu, Hainiu and Ippolito,
                   Daphne and Callison-Burch, Chris",
  month         =  may,
  year          =  2024,
  archivePrefix = "arXiv",
  primaryClass  = "cs.CL",
  eprint        = "2405.07940"
}

@ARTICLE{Verma2023-ti,
  title         = "Ghostbuster: Detecting Text Ghostwritten by Large Language
                   Models",
  author        = "Verma, Vivek and Fleisig, Eve and Tomlin, Nicholas and
                   Klein, Dan",
  month         =  may,
  year          =  2023,
  archivePrefix = "arXiv",
  primaryClass  = "cs.CL",
  eprint        = "2305.15047"
}

@ARTICLE{Latona2024-lw,
  title         = "The {AI} Review Lottery: Widespread {AI-Assisted} Peer
                   Reviews Boost Paper Scores and Acceptance Rates",
  author        = "Latona, Giuseppe Russo and Ribeiro, Manoel Horta and
                   Davidson, Tim R and Veselovsky, Veniamin and West, Robert",
  month         =  may,
  year          =  2024,
  archivePrefix = "arXiv",
  primaryClass  = "cs.CY",
  eprint        = "2405.02150"
}

@ARTICLE{Kiela2021-tx,
  title         = "Dynabench: Rethinking Benchmarking in {NLP}",
  author        = "Kiela, Douwe and Bartolo, Max and Nie, Yixin and Kaushik,
                   Divyansh and Geiger, Atticus and Wu, Zhengxuan and Vidgen,
                   Bertie and Prasad, Grusha and Singh, Amanpreet and Ringshia,
                   Pratik and Ma, Zhiyi and Thrush, Tristan and Riedel,
                   Sebastian and Waseem, Zeerak and Stenetorp, Pontus and Jia,
                   Robin and Bansal, Mohit and Potts, Christopher and Williams,
                   Adina",
  month         =  apr,
  year          =  2021,
  archivePrefix = "arXiv",
  primaryClass  = "cs.CL",
  eprint        = "2104.14337"
}

@INPROCEEDINGS{Zhang2020-tf,
  title     = "Effect of confidence and explanation on accuracy and trust
               calibration in {AI-assisted} decision making",
  booktitle = "Proceedings of the 2020 Conference on Fairness, Accountability,
               and Transparency",
  author    = "Zhang, Yunfeng and Liao, Q Vera and Bellamy, Rachel K E",
  publisher = "Association for Computing Machinery",
  pages     = "295--305",
  series    = "FAT* '20",
  month     =  jan,
  year      =  2020,
  address   = "New York, NY, USA",
  location  = "Barcelona, Spain"
}

@ARTICLE{Atanasova2020-uz,
  title         = "A diagnostic study of explainability techniques for text
                   classification",
  author        = "Atanasova, Pepa and Simonsen, Jakob Grue and Lioma, Christina
                   and Augenstein, Isabelle",
  journal       = "arXiv [cs.CL]",
  month         =  sep,
  year          =  2020,
  archivePrefix = "arXiv",
  primaryClass  = "cs.CL"
}

@ARTICLE{Hu2023-uv,
  title         = "{RADAR}: Robust {AI}-Text Detection via Adversarial Learning",
  author        = "Hu, Xiaomeng and Chen, Pin-Yu and Ho, Tsung-Yi",
  journal       = "arXiv [cs.CL]",
  month         =  jul,
  year          =  2023,
  archivePrefix = "arXiv",
  primaryClass  = "cs.CL"
}

@article{krishna2023paraphrasing,
  title={Paraphrasing evades detectors of ai-generated text, but retrieval is an effective defense},
  author={Krishna, Kalpesh and Song, Yixiao and Karpinska, Marzena and Wieting, John and Iyyer, Mohit},
  journal={Advances in Neural Information Processing Systems},
  volume={36},
  pages={27469--27500},
  year={2023}
}

@ARTICLE{Dawkins2025-cg,
  title         = "When detection fails: The power of fine-tuned models to
                   generate human-like social media text",
  author        = "Dawkins, Hillary and Fraser, Kathleen C and Kiritchenko,
                   Svetlana",
  journal       = "arXiv [cs.CL]",
  month         =  jun,
  year          =  2025,
  archivePrefix = "arXiv",
  primaryClass  = "cs.CL"
}

@INPROCEEDINGS{Zhang2024-di,
  title     = "Detection Vs. Anti-detection: Is Text Generated by {AI}
               Detectable?",
  author    = "Zhang, Yuehan and Ma, Yongqiang and Liu, Jiawei and Liu,
               Xiaozhong and Wang, Xiaofeng and Lu, Wei",
  booktitle = "Wisdom, Well-Being, Win-Win",
  publisher = "Springer Nature Switzerland",
  pages     = "209--222",
  year      =  2024
}

@ARTICLE{Kadiyala2025-kc,
  title         = "Robust and fine-grained detection of {AI} generated texts",
  author        = "Kadiyala, Ram Mohan Rao and Pullakhandam, Siddartha and
                   Mehreen, Kanwal and Sharma, Drishti and Gupta, Siddhant and
                   Purbey, Jebish and Srivastava, Ashay and TippaReddy, Subhasya
                   and Bobbili, Arvind Reddy and Chandrashekhar, Suraj Telugara
                   and Adeeb, Modabbir and Vura, Srinadh and Farooq, Hamza",
  journal       = "arXiv [cs.CL]",
  month         =  apr,
  year          =  2025,
  archivePrefix = "arXiv",
  primaryClass  = "cs.CL"
}

@ARTICLE{Pham2025-fe,
  title         = "Frankentext: Stitching random text fragments into long-form
                   narratives",
  author        = "Pham, Chau Minh and Russell, Jenna and Pham, Dzung and Iyyer,
                   Mohit",
  journal       = "arXiv [cs.CL]",
  month         =  may,
  year          =  2025,
  archivePrefix = "arXiv",
  primaryClass  = "cs.CL"
}

@ARTICLE{Artemova2024-jh,
  title         = "Beemo: Benchmark of expert-edited machine-generated Outputs",
  author        = "Artemova, Ekaterina and Lucas, Jason and Venkatraman, Saranya
                   and Lee, Jooyoung and Tilga, Sergei and Uchendu, Adaku and
                   Mikhailov, Vladislav",
  journal       = "arXiv [cs.CL]",
  month         =  nov,
  year          =  2024,
  archivePrefix = "arXiv",
  primaryClass  = "cs.CL"
}

@ARTICLE{Ma2024-wz,
  title         = "Zero-shot detection of {LLM}-generated text using token
                   cohesiveness",
  author        = "Ma, Shixuan and Wang, Quan",
  journal       = "arXiv [cs.CL]",
  month         =  sep,
  year          =  2024,
  archivePrefix = "arXiv",
  primaryClass  = "cs.CL"
}

@ARTICLE{Zeng2024-qc,
  title         = "Detecting {AI}-generated sentences in human-{AI}
                   collaborative hybrid texts: Challenges, strategies, and
                   insights",
  author        = "Zeng, Zijie and Liu, Shiqi and Sha, Lele and Li, Zhuang and
                   Yang, Kaixun and Liu, Sannyuya and Gašević, Dragan and Chen,
                   Guanliang",
  journal       = "arXiv [cs.CL]",
  month         =  mar,
  year          =  2024,
  archivePrefix = "arXiv",
  primaryClass  = "cs.CL"
}

@INPROCEEDINGS{Richburg2024-al,
  title     = "Automatic Authorship Analysis in Human-{AI} Collaborative Writing",
  author    = "Richburg, Aquia and Bao, Calvin and Carpuat, Marine",
  booktitle = "Proceedings of the 2024 Joint International Conference on
               Computational Linguistics, Language Resources and Evaluation
               (LREC-COLING 2024)",
  pages     = "1845--1855",
  year      =  2024
}

@ARTICLE{He2024-ox,
  title         = "Can watermarks survive translation? On the cross-lingual
                   consistency of text watermark for large language models",
  author        = "He, Zhiwei and Zhou, Binglin and Hao, Hongkun and Liu, Aiwei
                   and Wang, Xing and Tu, Zhaopeng and Zhang, Zhuosheng and
                   Wang, Rui",
  journal       = "arXiv [cs.CL]",
  month         =  feb,
  year          =  2024,
  archivePrefix = "arXiv",
  primaryClass  = "cs.CL"
}

@ARTICLE{Emi2024-nq,
  title         = "Technical report on the pangram {AI}-generated text
                   classifier",
  author        = "Emi, Bradley and Spero, Max",
  journal       = "arXiv [cs.CL]",
  month         =  feb,
  year          =  2024,
  archivePrefix = "arXiv",
  primaryClass  = "cs.CL"
}

@ARTICLE{Guo2023-gu,
  title         = "How close is {ChatGPT} to human experts? Comparison corpus,
                   evaluation, and detection",
  author        = "Guo, Biyang and Zhang, Xin and Wang, Ziyuan and Jiang, Minqi
                   and Nie, Jinran and Ding, Yuxuan and Yue, Jianwei and Wu,
                   Yupeng",
  journal       = "arXiv [cs.CL]",
  month         =  jan,
  year          =  2023,
  archivePrefix = "arXiv",
  primaryClass  = "cs.CL"
}

@ARTICLE{Zhou2024-ud,
  title         = "Humanizing Machine-Generated Content: Evading {AI}-Text
                   Detection through Adversarial Attack",
  author        = "Zhou, Ying and He, Ben and Sun, Le",
  journal       = "arXiv [cs.CL]",
  month         =  apr,
  year          =  2024,
  archivePrefix = "arXiv",
  primaryClass  = "cs.CL"
}

@INPROCEEDINGS{Zheng2025-lq,
  title     = "\textit{TH-bench:} evaluating evading attacks via humanizing {AI}
               text on machine-generated text detectors",
  author    = "Zheng, Jingyi and Wang, Junfeng and Sun, Zhen and Dong, Wenhan
               and Liu, Yule and He, Xinlei",
  booktitle = "Proceedings of the 31st ACM SIGKDD Conference on Knowledge
               Discovery and Data Mining V.2",
  publisher = "ACM",
  address   = "New York, NY, USA",
  pages     = "5948--5959",
  month     =  aug,
  year      =  2025,
  language  = "en"
}

@ARTICLE{Chakrabarty2025-ke,
  title         = "Readers prefer outputs of {AI} trained on copyrighted books
                   over expert human writers",
  author        = "Chakrabarty, Tuhin and Ginsburg, Jane C and Dhillon,
                   Paramveer",
  journal       = "arXiv [cs.CL]",
  month         =  oct,
  year          =  2025,
  archivePrefix = "arXiv",
  primaryClass  = "cs.CL"
}

@inproceedings{karr2026academic,
  title={Why AI Detection Fails for Academic Integrity},
  author={Karr, Jonathan A. and Khvatskii, Grigorii and Hua, Ting and Chawla, Nitesh V.},
  booktitle={ACM AI Summit 2026},
  year={2026},
  doi={10.1145/3806096.3844871}
}

@inproceedings{saha2026peerreview,
  title={Policies Permitting LLM Use for Polishing Peer Reviews Are Currently Not Enforceable},
  author={Saha, Rounak and Juneja, Gurusha and Chaudhuri, Dayita and Sajeevan, Naveeja and Shah, Nihar B. and Pruthi, Danish},
  booktitle={International Conference on Machine Learning},
  year={2026}
}

@techreport{jabarian2025artificial,
  title={Artificial Writing and Automated Detection},
  author={Jabarian, Brian and Imas, Alex},
  institution={National Bureau of Economic Research},
  number={34223},
  year={2025},
  doi={10.3386/w34223}
}

@misc{lee2026epochdetectors,
  author={Lee, Jaeho},
  title={AI Detectors Rarely Flag Human Writing, but Sometimes Miss AI Text Imitating Real Authors},
  publisher={Epoch AI},
  year={2026},
  url={https://epoch.ai/data-insights/ai-detectors-false-negatives}
}

@article{li2026meld,
  title={MELD: Multi-Task Equilibrated Learning Detector for AI-Generated Text},
  author={Li, Chenjun and Wan, Cheng and Chen, Haomiao and Paetzold, Johannes C.},
  journal={arXiv preprint arXiv:2605.06903},
  year={2026}
}

@inproceedings{russell2025human,
  title={People Who Frequently Use ChatGPT for Writing Tasks Are Accurate and Robust Detectors of AI-Generated Text},
  author={Russell, Jenna and Karpinska, Marzena and Iyyer, Mohit},
  booktitle={Proceedings of ACL 2025},
  pages={5342--5373},
  year={2025},
  doi={10.18653/v1/2025.acl-long.267}
}

@article{vanvlasselaer2026who,
  title={Who Wrote This? Evaluating the Reliability of AI Detection Tools in Higher Education},
  author={Van Vlasselaer, Marijke and Van Droogenbroeck, Filip and Spruyt, Bram},
  journal={International Journal for Educational Integrity},
  volume={22},
  pages={16},
  year={2026},
  doi={10.1007/s40979-026-00226-w}
}

@inproceedings{Qazi_2024,
  series={WWW '24},
  title={GPT-generated Text Detection: Benchmark Dataset and Tensor-based Detection Method},
  url={http://dx.doi.org/10.1145/3589335.3651513},
  DOI={10.1145/3589335.3651513},
  booktitle={Companion Proceedings of the ACM Web Conference 2024},
  publisher={ACM},
  author={Qazi, Zubair and Shiao, William and Papalexakis, Evangelos E.},
  year={2024},
  month=may,
  pages={842--846},
  collection={WWW '24}
}

@InProceedings{pmlr-v306-duarte26a,
  title = 	 {Sem-Detect: Semantic Level Detection of {AI} Generated Peer-Reviews},
  author =       {Duarte, Andre Vicente and Tufts, Brian and Oke, Aditya and Fang, Fei and Oliveira, Arlindo L. and Li, Lei},
  booktitle = 	 {Proceedings of the 43rd International Conference on Machine Learning},
  pages = 	 {27089--27119},
  year = 	 {2026},
  editor = 	 {Zhang, Tong and Dudik, Miroslav and Jaggi, Martin and Agarwal, Alekh and Li, Sharon and Schuurmans, Dale and Zhu, Jerry and Berkenkamp, Felix and Dong, Hanze and Bietti, Alberto},
  volume = 	 {306},
  series = 	 {Proceedings of Machine Learning Research},
  month = 	 {06--11 Jul},
  publisher =    {PMLR},
  url = 	 {https://proceedings.mlr.press/v306/duarte26a.html}
}

@inproceedings{NEURIPS2024_b61bdf7e,
 author = {Wu, Junchao and Zhan, Runzhe and Wong, Derek and Yang, Shu and Yang, Xinyi and Yuan, Yulin and Chao, Lidia S.},
 booktitle = {Advances in Neural Information Processing Systems},
 doi = {10.52202/079017-3186},
 editor = {A. Globerson and L. Mackey and D. Belgrave and A. Fan and U. Paquet and J. Tomczak and C. Zhang},
 pages = {100369--100401},
 publisher = {Curran Associates, Inc.},
 title = {DetectRL: Benchmarking LLM-Generated Text Detection in Real-World Scenarios},
 url = {https://proceedings.neurips.cc/paper_files/paper/2024/file/b61bdf7e9f64c04ec75a26e781e2ad51-Paper-Datasets_and_Benchmarks_Track.pdf},
 volume = {37},
 year = {2024}
}

@misc{gehring2025gede,
  title         = {Assessing {LLM} Text Detection in Educational Contexts: Does Human Contribution Affect Detection?},
  author        = {Gehring, Lukas and Paa{\ss}en, Benjamin},
  year          = {2025},
  eprint        = {2508.08096},
  archivePrefix = {arXiv},
  primaryClass  = {cs.CL},
  note          = {Version 1. GEDESub is defined in v1 only},
  url           = {https://arxiv.org/abs/2508.08096v1}
}

@article{crossley2024persuade,
  title     = {A large-scale corpus for assessing written argumentation: {PERSUADE} 2.0},
  author    = {Crossley, Scott A. and Baffour, Perpetual and Tian, Yu and Franklin, Alex and Benner, Meg and Boser, Ulrich},
  journal   = {Assessing Writing},
  volume    = {61},
  pages     = {100865},
  year      = {2024},
  doi       = {10.1016/j.asw.2024.100865}
}

@article{naismith2022pelic,
  title     = {The University of Pittsburgh English Language Institute Corpus ({PELIC})},
  author    = {Naismith, Ben and Han, Na-Rae and Juffs, Alan},
  journal   = {International Journal of Learner Corpus Research},
  volume    = {8},
  number    = {1},
  pages     = {121--138},
  year      = {2022},
  doi       = {10.1075/ijlcr.21002.nai}
}

@article{leung2022cell,
  title     = {A learner corpus is born this way: From raw data to processed dataset},
  author    = {Leung, Chung Hong Danny and Chow, Mei Yung Vanliza and Ge, Haoyan},
  journal   = {Data in Brief},
  volume    = {44},
  pages     = {108527},
  year      = {2022},
  doi       = {10.1016/j.dib.2022.108527}
}

@article{crossley2023ellipse,
  title     = {The {English} Language Learner Insight, Proficiency and Skills Evaluation ({ELLIPSE}) Corpus},
  author    = {Crossley, Scott and Tian, Yu and Baffour, Perpetual and Franklin, Alex and Kim, Youngmeen and Morris, Wesley and Benner, Meg and Picou, Aigner and Boser, Ulrich},
  journal   = {International Journal of Learner Corpus Research},
  volume    = {9},
  number    = {2},
  pages     = {248--269},
  year      = {2023},
  doi       = {10.1075/ijlcr.22026.cro}
}

@misc{turnitin2023ell,
  title  = {New research: {Turnitin}'s {AI} detector shows no statistically significant bias against {English} language learners},
  author = {{Turnitin}},
  year   = {2023},
  url    = {https://www.turnitin.com/blog/new-research-turnitin-s-ai-detector-shows-no-statistically-significant-bias-against-english-language-learners},
  note   = {Blog post}
}

@misc{pangram2026esl,
  title  = {How accurate is {Pangram} {AI} detection on {ESL}?},
  author = {{Pangram Labs}},
  year   = {2026},
  url    = {https://www.pangram.com/blog/how-accurate-is-pangram-ai-detection-on-esl},
  note   = {Blog post, accessed November 2026}
}

@article{crossley2025asap,
  title     = {A large-scale corpus for assessing source-based writing quality: {ASAP} 2.0},
  author    = {Crossley, Scott A. and Baffour, Perpetual and Burleigh, Lauryn and King, Jules},
  journal   = {Assessing Writing},
  volume    = {65},
  pages     = {100954},
  year      = {2025},
  doi       = {10.1016/j.asw.2025.100954}
}

@book{ishikawa2023icnale,
  title     = {The {ICNALE} Guide: An Introduction to a Learner Corpus Study on {Asian} Learners' {L2} {English}},
  author    = {Ishikawa, Shin'ichiro},
  publisher = {Routledge},
  year      = {2023},
  doi       = {10.4324/9781003252528}
}

@article{liang2023gpt,
  title     = {{GPT} detectors are biased against non-native {English} writers},
  author    = {Liang, Weixin and Yuksekgonul, Mert and Mao, Yining and Wu, Eric and Zou, James},
  journal   = {Patterns},
  volume    = {4},
  number    = {7},
  pages     = {100779},
  year      = {2023},
  doi       = {10.1016/j.patter.2023.100779}
}
\newpage
\appendix

\section{Training Dataset Statistics}
\label{sec:training_data_stats}


Table~\ref{tab:training-data} shows the number of human written documents used during training split by domains and sub-domains. Every domain in the table also contains AI-generated and mixed training texts, which are
generated from or alongside its human texts covering a diverse set of LLMs and prompts. Having such a large dataset enables us to tune the data used at training time such that we can prioritize certain domains, or prioritize accuracy on human documents. 

\begin{table}[!htbp]
\centering
\small
\begin{tabular}{lrr}
\toprule
Domain & Human texts & Share (\%) \\
\midrule
Essays & 1.48M & 26.8 \\
\quad Expository essay & 1.08M & 19.5 \\
\quad Argumentative essay & 176K & 3.2 \\
\quad Descriptive essay & 125K & 2.3 \\
\quad Persuasive essay & 69.9K & 1.3 \\
\quad Admissions essay & 31.2K & 0.6 \\
News article & 796K & 14.4 \\
Blog post & 772K & 14.0 \\
Short story & 627K & 11.4 \\
Research paper & 418K & 7.6 \\
Encyclopedia article & 299K & 5.4 \\
Marketing copy & 249K & 4.5 \\
Press release & 170K & 3.1 \\
Legal brief & 158K & 2.9 \\
Product review & 131K & 2.4 \\
Social media post & 75.3K & 1.4 \\
Email & 59.7K & 1.1 \\
Poem & 57.3K & 1.0 \\
Service review & 54.5K & 1.0 \\
Restaurant review & 50.3K & 0.9 \\
Lab report & 37.7K & 0.7 \\
Financial disclosure & 20.6K & 0.4 \\
Forum post & 18.2K & 0.3 \\
Biography & 10.8K & 0.2 \\
Peer review & 8.1K & 0.1 \\
Transcript & 8.0K & 0.1 \\
About page & 6.8K & 0.1 \\
Speech & 6.6K & 0.1 \\
Resume/CV & 6.4K & 0.1 \\
\midrule
Total & 5.52M & 100.0 \\
\bottomrule
\end{tabular}
\caption{Human-written training texts of GPTZero 5.0b by writing domain, the data behind
Figure~\ref{fig:training-domains}.}
\label{tab:training-data}
\end{table}

\section{Challenges with Data}
Table \ref{table:dataset_bias} shows some of the challenges faced in terms of debiasing data to prevent the model from relying on spurious shortcuts.
\label{sec:}
\begin{table*}[h!]
    \centering
    \begin{tabular}{p{0.6\textwidth} p{0.2\textwidth} p{0.2\textwidth}}
        \toprule
         Example & Label & Dataset \\
        \midrule
        many lepidopteran insects are agricultural pests that affect stored grains , food and fiber crops . & Human & Scientific Papers \\
        \midrule
        The sine-Gordon field theory and its associated massive Thirring model are quantum field theories that have been extensively studied by researchers.  & AI & Scientific Papers \\
        \midrule
        Suddenly , Google could find itself in a position where it has to explain that they can not perform miracles to a fully committed audience invested in the idea that they can . All of these and many , many more ways to fall off exist . There 's no rest for the weary , as they say . :-) & Human & HC3 Plus \\
        \midrule
        These include the financial performance and prospects of the company, the strength of its management and leadership, the stability and growth potential of its industry, and the overall state of the economy.Additionally, the value of a share may be influenced by the demand for the stock among investors, as well as the supply of the stock that is available for purchase. & AI & HC3 Plus \\
        \bottomrule
    \end{tabular}
    \caption{Example of formatting biases. The human document in Scientific Papers is all lowercase with extraneous spacing around punctuation, while the AI document has correct capitalization and spacing. The human document in HC3 Plus has an emoji and extraneous spacing, while the AI document is more formal, and is missing space before "Additionally". }
    \label{table:dataset_bias}
\end{table*}

\section{Case Study Details}
\subsection{Baselines}
\label{app:baselines}
\textbf{Radar} The official Hugging Face implementation of Radar to compute a probability score. This is simply a RoBERTa model with a classification head \href{implementation}{https://huggingface.co/TrustSafeAI/RADAR-Vicuna-7B}. 

\noindent \textbf{Binoculars}
For Binoculars, we use Llama-2 7B and Llama-2 7B-chat as the two models used to compute the cross perplexity score. Input text is tokenized similarly for each model, and a sigmoid activation is used similar to perplexity to transform scores into a probability.

\noindent \textbf{Pangram} The Pangram API offers a `prediction\_short` field indicating the document-level prediction according to their internal threshold, while `fraction\_ai`, `fraction\_ai\_assisted`, `fraction\_human` indicate the percentage of the document that is entirely AI-generated, polished by AI, or written by a human respectively.
We use Pangram's V3 API in our analysis.

\noindent \textbf{Originality} The Originality API provides a `classification` field indicating the predicted class, and a `confidence` field indicating the likelihood that the predicted class is the true class. We use Originality's V3 API in our analysis.

\subsection{Datasets}
\label{sec:case_study_dataset_details}

The third-party benchmarks are described in Section~\ref{sec:third-party}, fairness corpora in Section~\ref{sec:fairness}, and the AI bypasser benchmark in Section~\ref{section:ai_bypassers} and Appendix~\ref{appendix:paraphraser_services}.
The domain and generating-model breakdowns (Tables~\ref{tab:domain-fpr}, \ref{tab:domain-recall} and \ref{tab:llms}) pool GPTZero's internal evaluation sets. These cover human-written web text, student and learner writing, and benchmarks of
human writing in individual domains; AI-generated text from frontier LLMs, including
generations from out-of-distribution prompts; AI text paraphrased or humanized by
paraphrasing models and commercial services; and native mixed documents in which an
LLM continues or fills in human text. GPTZero 5.0b
and GPTZero 4.11b are evaluated on identical samples and labels.

\section{Internal Evaluation Set}
\label{sec:internal_evaluation_set}
Beyond publicly available benchmarks, we use an internal evaluation set for evaluating our model. This evaluation set contains 40k documents with label proportion 40\% human, 40\% AI, and 20\% mixed. It is subsampled from the test split that was generated from splitting our training data into non-overlapping datasets, but it also includes out-of-distribution datasets for more challenging evaluation. 

\section{Formatting Benchmark Set}
\label{sec:formatting_benchmark_set}
We subset our Internal Evaluation Set (Section \ref{sec:internal_evaluation_set}) to approximately 2,000 documents which contain bold formatting, and another 2,000 documents which contain lists. For the bold formatting benchmark corresponding documents with the bold formatting removed were created, and for the list benchmark, documents with the list items merged into sentences were created. Both benchmarks are made up of roughly 60\% AI texts, and 40\% human texts.

\section{User Interface}
\begin{figure}[h!]
    \centering
    \includegraphics[width=\linewidth]{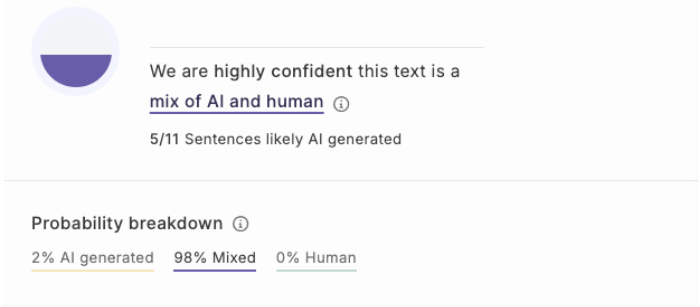}
    \caption{Main user interface for document scans. An overall Classification is presented along with a user-friendly confidence category. Per-class likelihoods are shown in the Probability Breakdown section.}
    \label{figure:user_interface}
\end{figure}

Figure \ref{figure:user_interface} shows the main \gptzero{} user interface, displaying the predicted class and associated confidence scores in the top \textit{Classification} section, along with tooltips explaining how to interpret these values. 
These explanations help ensure the responsible use of our detector, as many users lack background in statistics and hypothesis testing \cite{Zhang2020-tf}. 
Advanced users may also specify a confidence threshold appropriate for their required levels of precision and recall (see also Sec. \ref{sec:calibration}).
The \textit{Probability Breakdown} section then shows the probability assigned by our detector to the human, AI, and mixed classes, providing complementary information to the predicted class confidence.

\section{User Feedback}
Users are able to provide feedback on multiple aspects of our detector including accuracy and interface usability. 
\label{sec:user_feedback}


\section{Advanced Scan}
\label{sec:appendix_deep_scan}
\subsection{Visual example}

\begin{figure*}[h!]
    \centering
    \includegraphics[width=\linewidth]{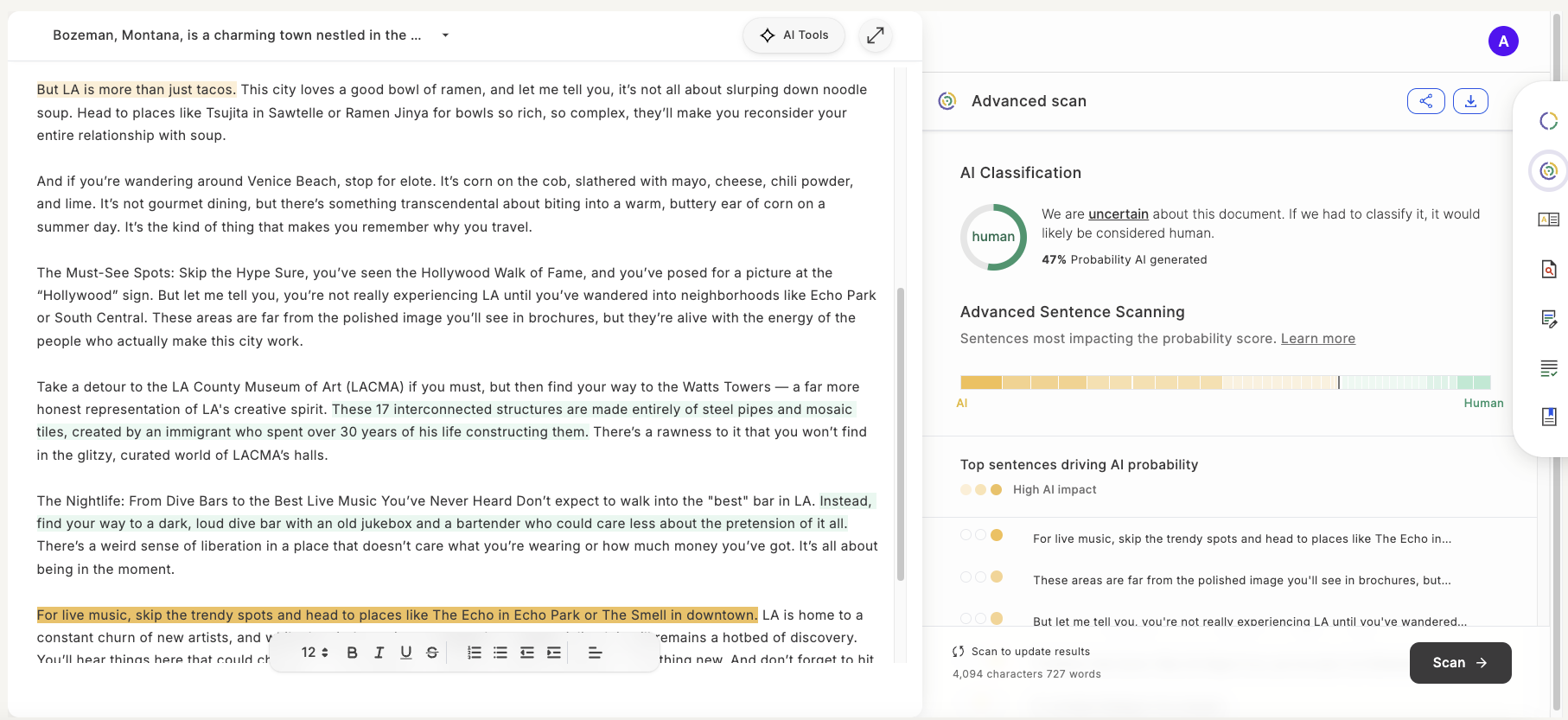}
    \caption{Example of the \gptzero{} Advanced Scan feature showing the most important sentences contributing to a prediction.}
    \label{fig:deep_scan}
\end{figure*}

Figure \ref{fig:deep_scan} visualizes our Advanced Scan feature on a GPT4-generated document. Scores indicate the importance of each sentence to the human or AI class. The document that was scanned has predicted to be AI with high confidence, but the second sentence in the second paragraph causes our detector to lower its predicted probability compared to what it otherwise would have been.

\section{Cloud Infrastructure}
Model deployments are hosted using Amazon Web Services (AWS) Elastic Container Services (ECS) on Nvidia Ampere GPUs.
We load test a single instance with 10 concurrent users sending requests at a constant rate and observe that each instance is capable of handling up to 10 requests per second with a median repsonse time of 480ms, and with 95\% of requests being served in under 860ms.

\begin{figure*}
    \centering
    \includegraphics[width=0.7\linewidth]{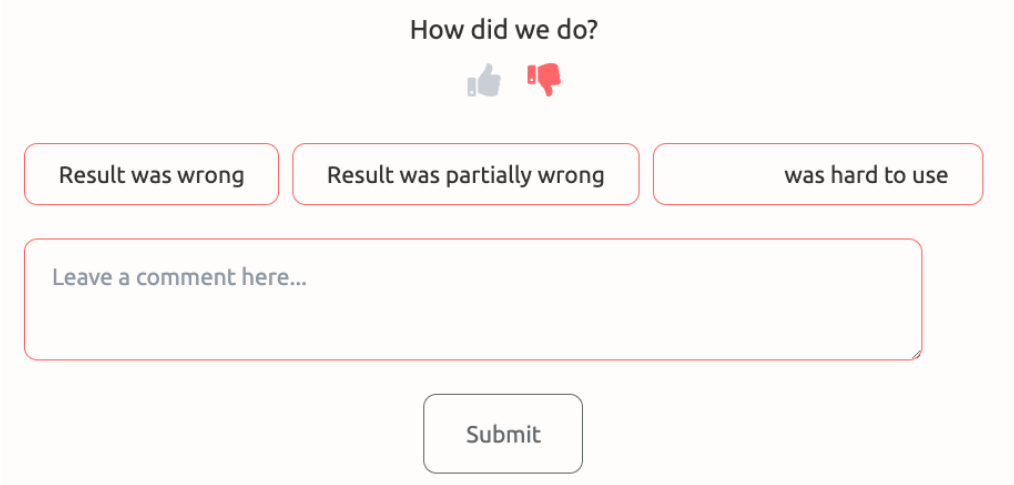}
    \caption{Our user feedback collection interface for improving predictions.}
    \label{figure:user_feedback}
\end{figure*}

\subsection{Classifier Details}
We trained an XGBoost classifier with AI and Human features that was applied to 2000 balanced AI and Human examples stratified across domains such as news, scientific writing, and creative writing and reviews where AI generations were obtained from a wide range of LLM model families such as openAI (GPT-4x, GPT-3.5), Claude and Gemini.

A grid-search was performed using the following parameters on a validation split and the best model selected for feature selection:

\begin{itemize}
    \item \textbf{Max Leaf Nodes}: [50, 100, 200, 300]
    \item \textbf{Max Depth}: [3, 5, 7, 10]
    \item \textbf{Number of Trees}: [100, 200, 300, 500]
    \item \textbf{Learning Rate}: [0.01, 0.05, 0.1, 0.2, 0.3]
\end{itemize}

Using the resultant model, we reduced an initial feature set comprising $\sim 200$ features per label to $\sim 20$ based on feature importances.


\subsection{Classification Results}
\begin{table}[H]
    \centering
    \resizebox{0.9\columnwidth}{!}{
        \begin{tabular}{lrrrr}
        \toprule
             Class & Precision (\%) & Recall (\%) & Accuracy (\%) \\
        \midrule
             AI    & 90.0 & 91.0 & 91.0 \\
             Human & 91.0 & 90.0 & 90.0 \\
        \bottomrule
        \end{tabular}
    }
    \caption{Classification metrics (in \%) for AI and Human classes using mined features.}
    \label{table:classification_metrics_nlu}
\end{table}

\newpage
\section{Sub-class separability}

\begin{figure}[h!]
\centering
\includegraphics[scale=0.4]{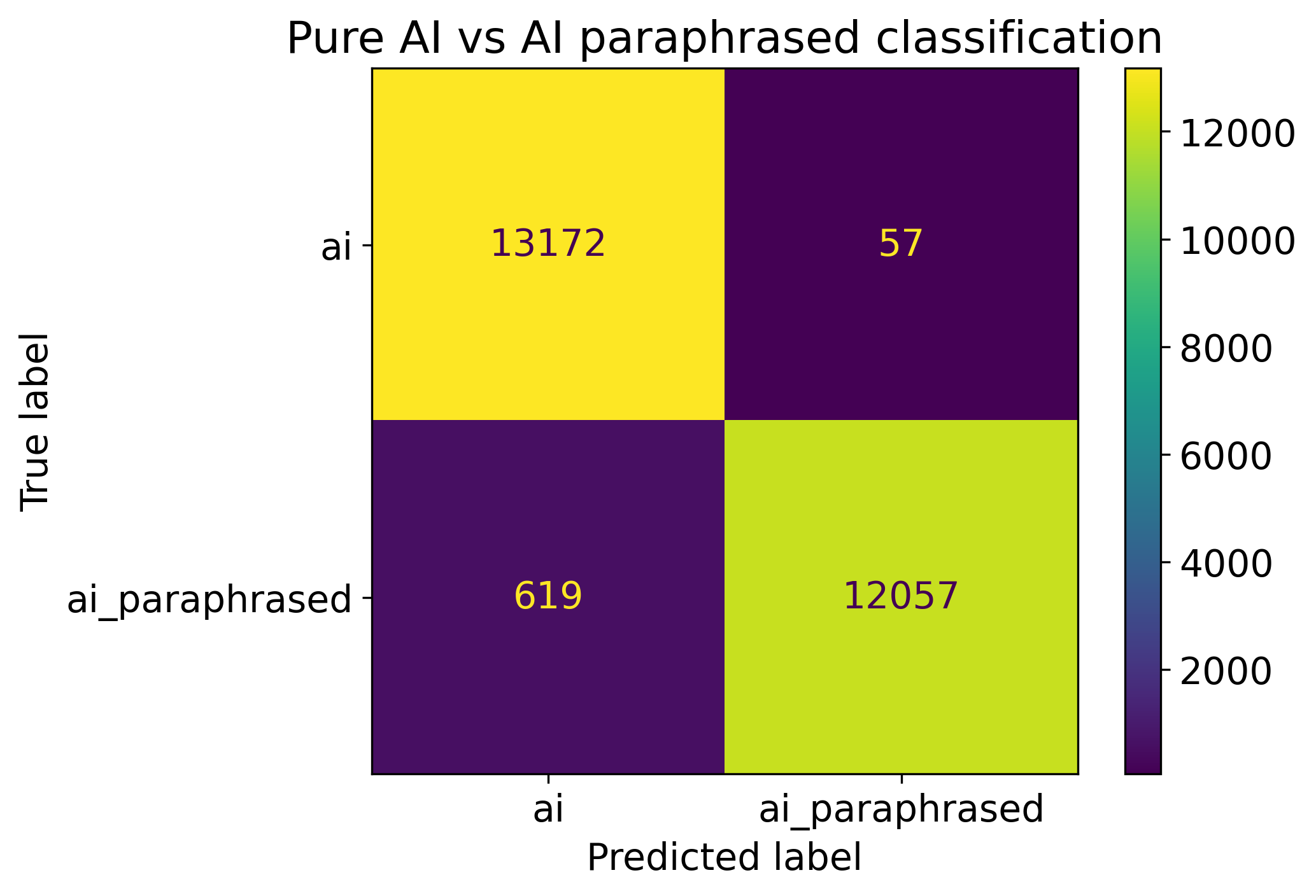}
\caption{Confusion Matrix representing AI vs Pure AI classes.}
\label{figure:confusion_matrix_aip_pureai}
\end{figure}

We prepared a nearly balanced dataset consisting of paired AI and AI Paraphrased examples from multiple domains such as social media, consumer reviews, news, online articles/blogs, encyclopedia, student essays etc. and used it to benchmark our model. 
The sub-classes AI and AI-paraphrased are quite separable using our hierarchical classification methodology as seen in Fig. \ref{figure:confusion_matrix_aip_pureai}.

We note that we are able to achieve a FPR <=0.5\% on the AI class. However, we observe a higher number of False Negatives for the AI paraphrased class. This stems from the inclusion of outputs from lower quality AI-paraphraser tools and settings, which introduces minimal perturbations and are considered poor AI bypass attempts. Classifying such samples as LLM generated is an acceptable detection standard.

\section{Postprocessing Predictions to Limit False Positives}
\label{sec:calibration}
\begin{figure}[h]
    \centering
    \begin{subfigure}[b]{0.45\columnwidth}
        \centering
        \includegraphics[width=\textwidth]{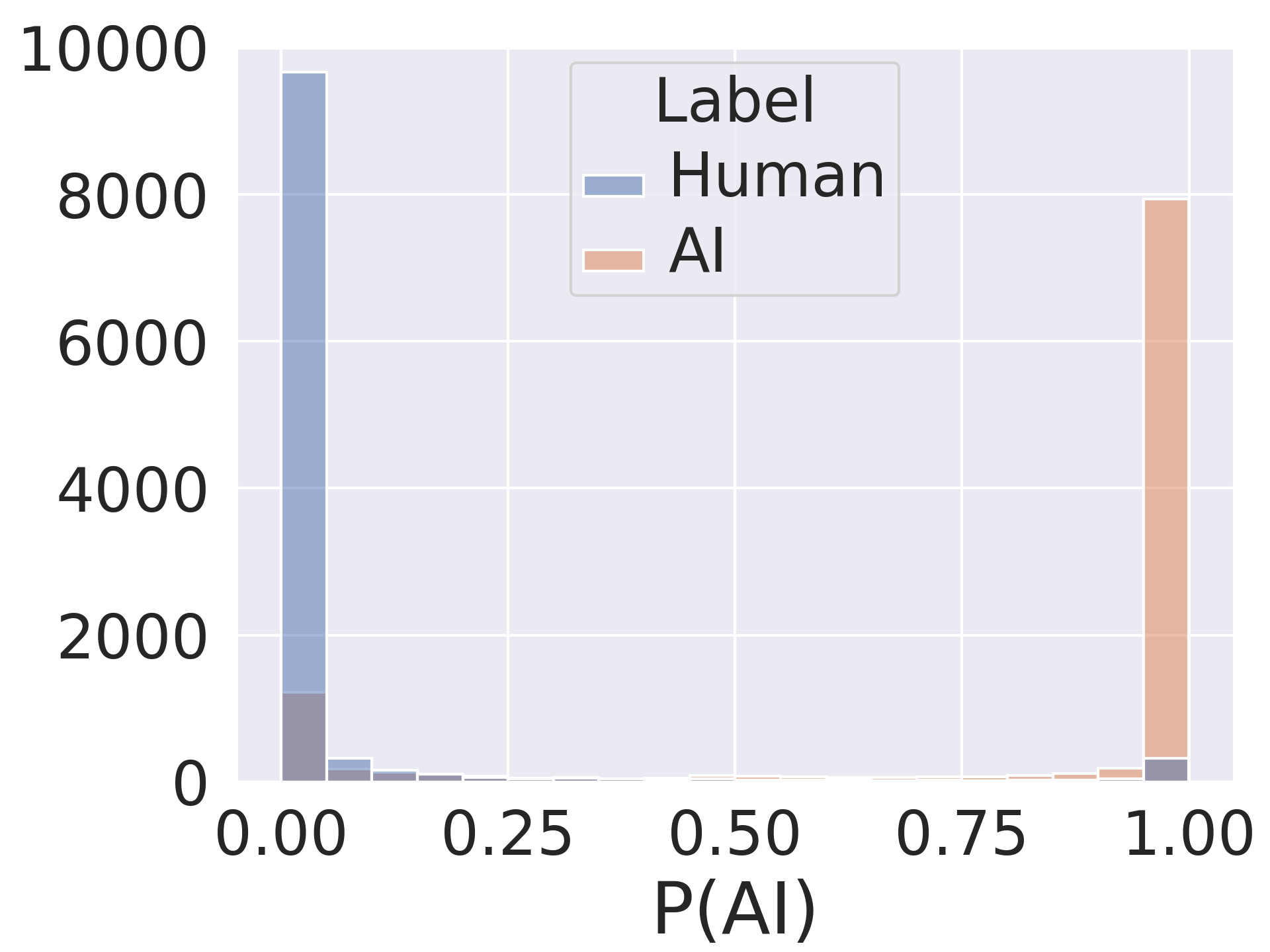}
        \caption{Output distribution without mapping (ECE=0.074)}
    \end{subfigure}
    \hfill
    \begin{subfigure}[b]{0.45\columnwidth}
        \centering
        \includegraphics[width=\textwidth]{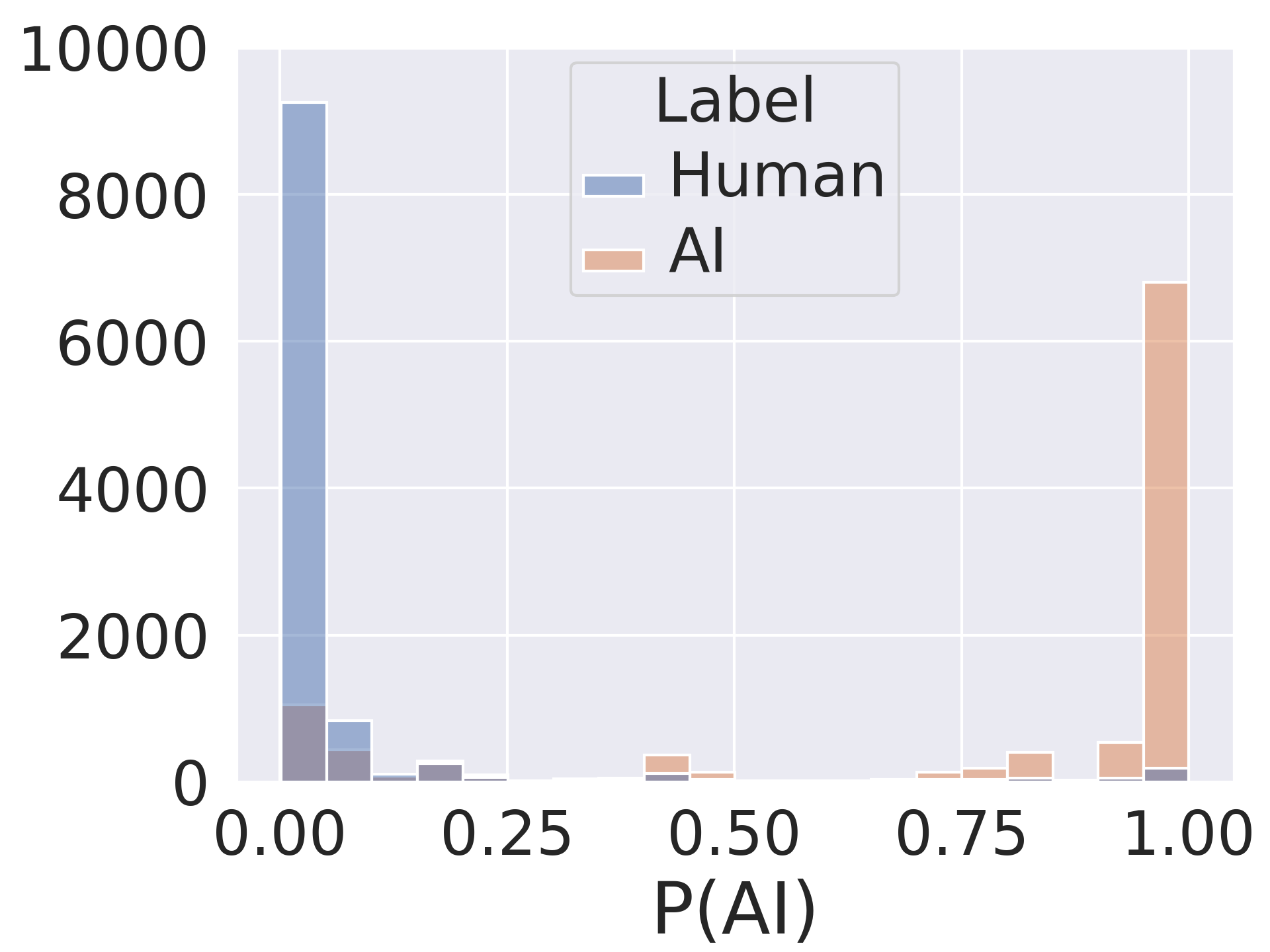}
        \caption{Output distribution with mapping (ECE=0.070)}
    \end{subfigure}
    \caption{Remapping outputs to reduce overconfident predictions limits the negative effects of false positives.}
    \label{figure:calibration}
\end{figure}

We consider it to be particularly important to limit false positive errors by transforming low confidence AI/mixed predictions to human. 
This is because most users will not specify thresholds of their own and false allegations of text being AI-written can have harmful consequences for the writer. 
Figure \ref{figure:calibration} compares our detector's output distribution on an internal benchmark of ~27,000 samples with and without our custom output mapping, and the expected calibration error (ECE) \cite{Guo2017-wo}.
\gptzero{}'s custom output mapping is optimized to both reduce ECE with a focus on penalizing overly confident predictions, while also reducing the most harmful effects of false positives. 

\begin{equation*}
    \mathrm{ECE} = \sum_{m=1}^{M} \frac{|B_{m}|}{n} |\mathrm{acc}(B_{m}) - \mathrm{conf}(B_{m})|
\end{equation*}
\noindent where $B_{m}=\{s\in\mathcal S:\mathrm{conf}(s)\in(\frac{m-1}{M},\frac{m}{M}]\}$ is formed as a bin of samples with similar confidence value, grouped into a total of $M$ bins. 
$\mathrm{acc}(B_{m})$ and $\mathrm{conf}(B_{m})$ are the average accuracy and confidence of samples in $B_{m}$. 

\section{Polishing Prompts}
\label{appendix:polishing_prompts}
Table \ref{table:polish_prompts} shows several examples of polishing prompts that we use to generate Polished texts, as well as example outputs for a given human document.

Human Document: 

\begin{table*}[t]
\centering
\small
\begin{tabular}{p{0.2\textwidth} p{0.8\textwidth}}
\toprule
\textbf{Prompt} & \textbf{Polished Text} \\
Original Human Document & The author gave a loving mood in the memoir. I know this becacus he did not talk badily about anything and he talked about how his parents helped him and other people as a kit. in paragraph two he was talking about how his house was nice and comsortabur. I know this because he sais. the love of cooking and the passioniate Cuban music. \\
Polish this text: & The author creates a warm, loving tone in the memoir. I know this because he doesn’t speak negatively about anything; instead, he focuses on how his parents helped him and others when he was a child. In paragraph two, he describes his home as nice and comfortable. I can tell because he mentions “the love of cooking” and “the passionate Cuban music,” which help paint a cozy, caring atmosphere.  \\
\midrule
Make this text sound more fluid: & The author creates a warm, loving tone throughout the memoir. I can tell because he doesn’t speak negatively about anything—instead, he focuses on how his parents supported him and cared for others when he was a child. In paragraph two, he describes his home as nice and comfortable, shown through details like the love of cooking and the passionate Cuban music. \\
\midrule
Improve the quality of this text while maintaining a human voice: & The author creates a warm, loving tone throughout the memoir. I can tell because he doesn’t speak negatively about anything. Instead, he focuses on how his parents supported him and how other people helped him when he was a kid. In paragraph two, he describes his home as nice and comfortable. You can see this in the details he shares, like the love of cooking and the passionate Cuban music. \\
\midrule
Improve the spelling and grammar in this text: & The author created a loving mood in the memoir. I know this because he did not speak badly about anything. He talked about how his parents helped him and other people when he was a kid. In paragraph two, he talked about how his house was nice and comfortable. I know this because he says he loved cooking and passionate Cuban music.\\
\bottomrule
\end{tabular}
\caption{A comparison of some polishing prompts used for generated texts trained under the Polished class.}
\label{table:polish_prompts}
\end{table*}

\section{Black-Box Paraphraser Services}
\label{appendix:paraphraser_services}
Table~\ref{table:paraphraser_services} lists the commercial humanizer and paraphrasing
services used to create the bypassed AI texts in our AI bypasser evaluation.

\begin{table}[!htbp]
\centering
\small
\begin{tabular}{ll}
\toprule
\textbf{Service} & \textbf{Bypassing ability on naive detectors} \\
\midrule
GPTinf         & Low    \\
Grubby AI      & Medium \\
HIX            & Low    \\
HumanizeAIText & Medium \\
Phrasely       & High   \\
QuillBot       & Low    \\
RyanAI         & Medium \\
StealthWriter  & Low    \\
TwainGPT       & Medium \\
Undetectable   & High   \\
WriteHuman     & Medium \\
\bottomrule
\end{tabular}
\caption{Commercial humanizer and paraphrasing services used to create the bypassed AI
texts in the AI bypasser evaluation (Section~5.5), with their bypassing ability on naive
AI detection methods.}
\label{table:paraphraser_services}
\end{table}

\end{document}